\documentclass[journal, twoside]{IEEEtran}
\ifCLASSINFOpdf
  \usepackage[pdftex]{graphicx}
\else
\fi
\usepackage{amsmath}
\usepackage{algorithm}
\usepackage{algorithmic}

\usepackage{array}

\usepackage[caption=false,font=footnotesize]{subfig}
\usepackage{comment}
\usepackage{color}
\usepackage{xr}
\usepackage{bm}
\begin{document}
%
\title{Experience-Based Evolutionary Algorithms for Expensive Optimization}

\author{Xunzhao~Yu, 
        Yan~Wang,
        Ling~Zhu,
        Dimitar~Filev,~\IEEEmembership{Fellow,~IEEE}
        and~Xin~Yao,~\IEEEmembership{Fellow,~IEEE}
\thanks{This work has been submitted to the IEEE for possible publication. Copyright may be transferred without notice, after which this version may no longer be accessible.}%
\thanks{This work was supported by Ford USA through a research grant to the University of Birmingham, UK. 
Xin Yao was also supported by the European Union's Horizon 2020 research and innovation programme under grant agreement number 766186 (ECOLE),  the Guangdong Provincial Key Laboratory (Grant No. 2020B121201001), the Program for Guangdong Introducing Innovative and Enterpreneurial Teams (Grant No. 2017ZT07X386), and Shenzhen Science and Technology Program (Grant No. KQTD2016112514355531).}
\thanks{X. Yu and X. Yao (the corresponding author) are with the CERCIA, 
	School of Computer Science, 
	University of Birmingham, Birmingham, B15 2TT, U.K. 
	(e-mail: xxy653@cs.bham.ac.uk; x.yao@cs.bham.ac.uk).
	X. Yao is also with the Research Institute of Trustworthy Autonomous Systems, and Guangdong Provincial Key Laboratory of Brain-inspired Computation, Department of Computer Science and Engineering, Southern University of Science and Technology, Shenzhen, China.}
\thanks{Y. Wang, L. Zhu, and D. Filev are with the Ford Motor Company, 
	2101 Village Rd, Dearborn, MI 48124, USA. 
	(e-mail: ywang21@ford.com, lzhu40@ford.com, dfilev@ford.com)}
}

\markboth{}
{Yu \MakeLowercase{\textit{et al.}}: Experience-based Evolutionary Algorithms for Expensive Optimization}

\maketitle

\begin{abstract} 
	Optimization algorithms are very different from human optimizers. A human being would gain more experiences through problem-solving, which helps her/him in solving a new unseen problem. Yet an optimization algorithm never gains any experiences by solving more problems. 
	In recent years, efforts have been made towards endowing optimization algorithms with some abilities of experience learning, which is regarded as experience-based optimization. 
	In this paper, we argue that hard optimization problems could be tackled efficiently by making better use of experiences gained in related problems. 
	We demonstrate our ideas in the context of expensive optimization, where we aim to find a near-optimal solution to an expensive optimization problem with as few fitness evaluations as possible. 	
	To achieve this, we propose an experience-based surrogate-assisted evolutionary algorithm (SAEA) framework to enhance the optimization efficiency of expensive problems, where experiences are gained across related expensive tasks via a novel meta-learning method. These experiences serve as the task-independent parameters of a deep kernel learning surrogate, then the solutions sampled from the target task are used to adapt task-specific parameters for the surrogate.
	With the help of experience learning, competitive regression-based surrogates can be initialized using only 1$d$ solutions from the target task ($d$ is the dimension of the decision space).
	Our experimental results on expensive multi-objective and constrained optimization problems demonstrate that experiences gained from related tasks are beneficial for the saving of evaluation budgets on the target problem. 
\end{abstract}


\IEEEpeerreviewmaketitle

\section{Introduction}\label{sec: introduction} 
	\IEEEPARstart{W}{hen} solving a new unseen problem, a human being always benefits from experiences gained from related problems he/she has seen in the past. Such an ability of experience learning enhances the efficiency of solving new problems and thus is desirable for intelligent optimization algorithms. 
	To endow optimization algorithms with some abilities of experience learning, many efforts have been made to combine diverse experience learning techniques with optimization algorithms, resulting in experience-based optimization algorithms \cite{Liu:2017:EBO, Tang:2021:FSP, Tan:2021:ETO}. 
	In the past decade, experience-based optimization approaches such as evolutionary transfer optimization \cite{Tan:2021:ETO, Jiang:2020:AFD, Jiang:2020:IBT}  and evolutionary multi-tasking optimization (EMTO) \cite{Wei:2021:MTR, Bali:2019:MFEAII, Xue:2020:ATE} have been proposed to solve diverse optimization problems, including automatic parameter tuning problems \cite{Tang:2021:FSP}, dynamic optimization problems \cite{Ruan:2019:SSCI, Ruan:2020:CEC}. These studies have demonstrated that the idea of experience-based optimization is effective when solving hard optimization problems, experiences gained from past problems can be helpful for solving new unseen optimization problems.	
	
	The motivation of this paper is to demonstrate that the idea of experience-based optimization is also working for diverse expensive optimization problems. 
	In real-world applications, many expensive optimization problems are related since they are working on similar issues in the same domain. 
	For example, in the domain of gasoline engine calibration, many calibration problems can be treated as related tasks. Although the categories of gasoline engines to be calibrated are different, the physical properties of gasoline will not change. Moreover, the mechanical structures of some gasoline engines are similar \cite{Yu:2022:ECS}. These domain-specific features make many calibration problems related to each other. 
	It is desirable to explore the domain-specific features from related tasks and then use them as experiences in SAEAs. The learned experiences can help SAEAs to build reliable surrogates with a lower cost of fitness evaluations than before (e.g., 1$d$ evaluations). Consequently, these experience-based SAEAs can save more fitness evaluations (in surrogate initialization) than non-experience-based SAEAs and thus are more suitable for very expensive optimization problems, where only 150 or fewer fitness evaluations are allowed during the optimization.

	The related tasks considered in this paper are expensive and each of them can provide only a small dataset of evaluated samples for experience learning. Therefore, our experience-based SAEA is based on the context of few-shot problems \cite{Chen:2019:Closer, Wang:2020:GFA}, where plenty of small related tasks are available for experience learning. 
	A challenge is that most existing experience-based optimization approaches can not learn experiences from small related tasks. However, recently, meta-learning \cite{Hospedales:2021:Survey} has been proved to be powerful in solving few-shot problems. 
	In meta-learning, the underlying common features of related tasks are extracted as past experiences, which can be integrated with the solutions sampled from new tasks and thus enhance the learning efficiency for new tasks.
	Quite a few meta-learning methods \cite{Wang:2020:GFA} have been proposed for few-shot problems. Benefiting from the ability of experience learning, these methods are capable of fitting accurate classification or regression models with limited samples from the target task, which motivates us to develop meta-learning methods to learn experiences for SAEAs.

	In this paper, we propose an experience-based SAEA framework to solve expensive optimization problems, including multi-objective optimization and constrained optimization.
	Major contributions are summarized as follows.
\begin{itemize}
	\item A novel meta-learning method is developed to gain experiences from related expensive tasks. 
	Based on the learned experiences, a regression-based surrogate is generated and then adapted to approximate the fitness landscape of the target task.
		The surrogate is derived from the deep kernel learning framework in which a Gaussian process employs a deep kernel to work as its covariance function. 
		Different from existing deep kernel learning models that are trained through meta-learning, our method learns only one common neural network for all related tasks and adapts task-specific parameters for the base kernel of a Gaussian process. Such a framework simplifies the architecture of experience learning as the model complexity will not grow with the number of past tasks, yet it is still able to adapt itself to a new task explicitly. 
	\item We propose an experience-based SAEA framework to combine regression-based SAEAs with our experience learning method. The framework employs our meta-learning model as surrogates for experience learning, and an update strategy is designed to adapt surrogates in each generation. 
	Note that our SAEA framework is a general framework as it is not designed for any specific scenario of optimization problems. 
	However, due to the page limitation, this paper focuses on only the scenarios of multi-objective optimization and constrained optimization, which have not been investigated before.
	\item We provide empirical evidence to show the effectiveness of our experience-based SAEA framework in expensive multi-objective optimization and expensive constrained single-objective optimization.
	Our ablation studies also discover the influence of some factors on the performance of experience-based optimization. 
\end{itemize}
 
The rest of this paper is organized as follows:
Section \ref{sec: background} introduces the related work and some preliminaries.
The details of the proposed method are described in Section \ref{sec: algorithm}. 
Section \ref{sec: experiment} presents experimental studies and analysis.
The conclusions and some further work are given in Section \ref{sec: conclusion}.

\section{Background} \label{sec: background} 
\subsection{Related Work}
In the past decade, experience-based evolutionary optimization has attracted much attention as it uses experiences gained from other optimization problems to improve the optimization efficiency of target problems, which mimics human capabilities of cognitive and knowledge generalization \cite{Gupta:2017:TOS}. 
The optimization problems that provide experiences or knowledge are regarded as source tasks, while the target optimization problems are regarded as target tasks.
To obtain useful experiences, the tasks that are related to target tasks are chosen as source tasks since they usually share domain-specific features with target tasks. 
Diverse experience-based evolutionary optimization methods have been proposed to use experiences gained from related tasks to tackle target tasks. They can be divided into two categories based on the direction of experience transformation. 

In the first category, experiences are transformed mutually. Every considered optimization problem is a target task and also is one of the source tasks of other optimization problems. In other words, the roles of source task and target task are compatible.
One representative tributary is EMTO that aims to solve multiple optimization tasks concurrently \cite{Ding:2017:GMF, Wei:2021:MTR, Liaw:2019:EMO, Bali:2019:MFEAII, Xue:2020:ATE}. 
In EMTO, experiences are learned, updated, and spontaneously shared among target tasks through multi-task learning techniques. 
A variant of EMTO is multiforms optimization \cite{Gupta:2017:TOS, Zhang:2021:ASO, Guo:2022:GMBO}. In multiforms optimization, multi-task learning methods are employed to learn experiences from distinct formulations of a single target task.

In the second category, experiences are transformed unidirectionally. The roles of source task and target task are not compatible, an optimization problem cannot serve as a source task and a target task simultaneously. 
One popular tributary is transfer optimization which employs transfer learning techniques to transform experiences from source tasks to target tasks \cite{Tan:2021:ETO, Jiang:2020:AFD, Jiang:2020:IBT, Volpp:2020:MetaBO}. 
In transfer learning, experiences can be transformed from a single source task, multiple source tasks, or even source tasks from a different domain \cite{Zhuang:2020:Review-TL}.
However, these transfer learning techniques pay more attention to experience transformation instead of experience learning. Despite diverse and complex situations of experience transformation have been studied \cite{Ruan:2019:SSCI, Ruan:2020:CEC}, the difficult of learning experiences from small (expensive) source tasks has not been well studied.
Actually, a common scenario in transfer learning is that the source task(s) is/are large enough such that useful experiences can be obtained easily through solving source task(s) \cite{Zhuang:2020:Review-TL}.
In contrast to transfer optimization, recently, some experience-based optimization algorithms attempted to use meta-learning methods to learn experiences from small source tasks, which are known as few-shot optimization (FSO)\cite{Wistuba:2021:FSBO}. Since meta-learning only works for related tasks in the same domain, the situations of experience transformation are less complex than that of transfer learning. As a result, meta-learning pays more attention to experience learning instead of experience transformation. Domain-specific features are extracted as experiences and no related task needs to be solved. 

	Our work belongs to the FSO in the second category discussed above since the experiences are transformed unidirectionally. More importantly, our experiences are learned across many related expensive tasks, rather than gained through solving more or less source tasks.
Existing studies on FSO only work for global optimization \cite{Wistuba:2021:FSBO}, leaving other optimization scenarios such as multi-objective optimization and constrained optimization still awaiting for investigation. 
In addition, in-depth ablation studies are lacking in the literature, making it unclear which factors affect the performance of FSO. Moreover, some studies use existing meta-learning models \cite{Patacchiola:2020:DKT} as their surrogates. No further adaptations are made to these surrogates during optimization since they were not originally designed for optimization. Therefore, a novel meta-learning model for FSO is desirable. 
	Our work fills the aforesaid gaps by proposing a novel meta-learning modeling method and a general experience-based SAEA framework. As a result, accurate surrogates and competitive optimization results can be achieved while the cost of surrogate initialization is only 1$d$ evaluations on the expensive target problem.

\subsection{Preliminaries} 
A meta deep kernel learning modeling method is developed to learn experiences in our SAEA framework. This subsection gives preliminaries about meta-learning and deep kernel learning. The former is the method of experience learning, the latter is the underlying structure of experience representation.
\subsubsection{Meta-Learning in Few-Shot Problems}
	In the context of few-shot problems, we have plenty of related tasks, each task $T$ contributes a couple of small datasets $D = \{(S, Q)\}$, namely support dataset $S$ and query dataset $Q$, respectively. After learning from datasets of random related tasks, a support set $S_*$ from new unseen task $T_*$ is given and one is asked to estimate the labels or values of a query set $Q_*$. The problem is called 1-shot or 5-shot when only 1 data point or 5 data points are provided in $S_*$. A comprehensive definition of few-shot problems is available in \cite{Chen:2019:Closer, Wang:2020:GFA}.
	
	Meta-learning methods have been widely used to solve few-shot problems \cite{Wang:2020:GFA}. They learn domain-specific features that are shared among related tasks as experiences, these experiences are used to understand and interpret the data collected from new tasks encountered in the future. 

\subsubsection{Deep Kernel Learning}
Deep kernel learning (DKL) \cite{Wilson:2016:DKL} aims at constructing kernels that encapsulate the expressive power of deep architectures for Gaussian processes (GPs \cite{Rasmussen:2006:GP}, also known as the Kriging model \cite{Stein:1999:Kriging} or the design and analysis of computer experiments (DACE) stochastic process model \cite{Sacks:1989:DACE}). To create expressive and scalable closed form covariance kernels, DKL combines the non-parametric flexibility of kernel methods and the structural properties of deep neural networks. 
In practice, a deep kernel $k(\textbf{x}^i, \textbf{x}^j | \bm{\gamma})$ transforms the inputs $\textbf{x}$ of a base kernel $k(\textbf{x}^i, \textbf{x}^j |\bm{\theta})$ through a non-linear mapping given by a deep architecture $\phi(\textbf{x} |\textbf{w}, \textbf{b})$:
\begin{equation} \label{equ: dk}
	k(\textbf{x}^i, \textbf{x}^j | \bm{\gamma}) = k( \phi(\textbf{x}^i | \textbf{w}, \textbf{b}), \phi(\textbf{x}^j | \textbf{w}, \textbf{b}) | \bm{\theta}),
\end{equation}
where $\bm{\theta}$ and $(\textbf{w}, \textbf{b})$ are parameter vectors of the base kernel and the deep architecture, respectively. $\bm{\gamma} = \{\bm{\theta}, \textbf{w}, \textbf{b}\}$ is a set of all parameters in this deep kernel. Note that in DKL, all parameters $\bm{\gamma}$ of a deep kernel $k(\textbf{x}^i, \textbf{x}^j | \bm{\gamma})$ are learned jointly by using the log marginal likelihood function of GPs as a loss function. Such a jointly learning strategy makes a DKL algorithm outperforms a combination of a deep neural network and a GP model, where a trained GP model is applied to the output layer of a trained deep neural network \cite{Wilson:2016:DKL}. 

\subsubsection{Meta-Learning on DKL}
An important distinction between DKL algorithms and the applications of meta-learning on DKL is that DKL algorithms learn their deep kernels from single tasks instead of collections of related tasks. Such a difference alleviates two drawbacks of single task DKL \cite{Tossou:2019:ADKL}: 
\begin{itemize}
	\item First, the scalability of deep kernels is no longer an issue as each dataset in meta-learning is small. 
	\item Second, the risk of overfitting is decreased since diverse data points are sampled across tasks.
\end{itemize}

\section{Experience-Based SAEA Framework} \label{sec: algorithm} 
In this paper, expensive optimization is carried out in the context of few-shot problems \cite{Tang:2021:FSP}, where the expensive optimization problem to be solved is denoted as target task $T_*$, and plenty of small datasets $D_i$ sampled from related tasks $T_i$ are available for experience learning.
We propose a SAEA framework to learn experiences from $T_i$ and use experiences in $T_*$, which saves fitness evaluations for expensive optimization problems. 
A meta deep kernel learning (MDKL) modeling method is developed to learn experiences from $T_i$ and then initializes surrogates with a cost of 1$d$ evaluations on $T_*$. 
Our SAEA framework combines MDKL with existing regression-based SAEAs. A new update strategy is designed to maintain and adapt surrogates for SAEAs.
As a result, although much fewer evaluations from $T_*$ are used, our SAEA framework can still achieve competitive or even better optimization results than the SAEAs that are unable to learn experiences from $T_i$.

\subsection{Overall Working Mechanism} 
\begin{figure*}[!t]
	\centering
	\includegraphics[width=7.0in]{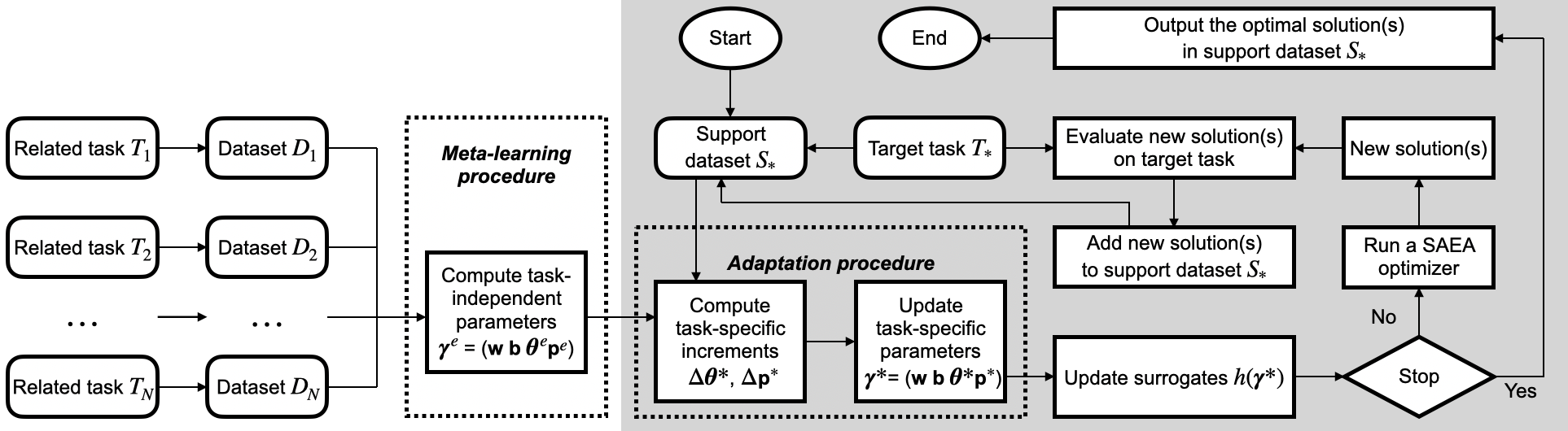}
	\caption[Diagram of our experience-based SAEA framework.]
	{Diagram of our experience-based SAEA framework.
	The grey block includes all modules that are related to the evolutionary optimization of target task $T_*$.
	The MDKL surrogate modeling method used in the framework consists of a meta-learning procedure and an adaptation procedure. The meta-learning procedure learns experiences (task-independent parameters $\bm{\gamma^e}$) from related tasks $T_i$. Based on the learned experiences, the adaptation procedure adapts MDKL task-specific parameters to approximate target task $T_*$.
	Note that existing SAEAs train and update their surrogates on $S_*$ only, thus their workflows do not contain a meta-learning procedure and they can not gain experiences from related tasks.}
	\label{fig: diagram}
\end{figure*}

A diagram of our experience-based SAEA framework is illustrated in Fig. \ref{fig: diagram}. 
All modules about the evolutionary optimization of target task $T_*$ are included in a grey block. The modules beyond the grey block are associated with related tasks $T_i$ and experience learning, which are the main distinctions between our SAEA framework and existing SAEAs.
The MDKL surrogate modeling method used in our SAEA framework consists of two procedures: meta-learning procedure and adaptation procedure. The former learns experiences from $T_i$, the latter uses experiences to adapt surrogates to fit $T_*$. 

The framework of experience-based SAEA is depicted in Algorithm \ref{alg: Framework}, it consists of the following major steps.
\begin{algorithm}[!t]
	\caption{Experience-Based SAEA Framework.}\label{alg: Framework}
	\textbf{Input:} \\
	\hspace*{\algorithmicindent} $D_i$: Datasets collected from related tasks $T_i$, i=$\{1, \dots, N\}$; \\
	\hspace*{\algorithmicindent} $N_m$: Number of datasets $D_m$ for meta-learning; \\
	\hspace*{\algorithmicindent} $|D_m|$: Size of $D_m$. Note that $|D_m| \le |D_i|$; \\
	\hspace*{\algorithmicindent} $B$: Number of related tasks in a batch; \\
	\hspace*{\algorithmicindent} $\alpha, \beta$: Surrogate learning rates; \\
	\hspace*{\algorithmicindent} $T_*$: Target task; \\
	\hspace*{\algorithmicindent} $Opt$: A SAEA optimizer; \\ 
	\hspace*{\algorithmicindent} $FE_{max}$: Fitness evaluation budget. \\
	\textbf{Procedure:} 
	\begin{algorithmic}[1]
		\STATE Experiences $\bm{\gamma}^e \leftarrow$ Meta-learning($D_i, N_m, |D_m|, B, \alpha$). 
		\STATE $S_* \leftarrow $ Sampling 1$d$ solutions from $T_*$.
		\STATE $h(\bm{\gamma}^*) \leftarrow$ Adaptation($\bm{\gamma}^e, S_*, \beta$). $/^*$Initialize surrogate.$^*/$
		\STATE Set evaluation counter $FE=|S_*|$. 
		\WHILE {$FE < FE_{max}$} 
			\STATE Candidate solution(s) $\textbf{x}^* \leftarrow$ Surrogate-assisted optimization ($Opt, h(\bm{\gamma}^*)$)		
			\STATE $f(\textbf{x}^*) \leftarrow$ Evaluate $\textbf{x}^*$ on $T_*$. 
			\STATE $S_* \gets S_* \cup \{(\textbf{x}^*, f(\textbf{x}^*))\}$. 
			\STATE $h(\bm{\gamma}^*) \leftarrow$  Update($\bm{\gamma}^*, S_*, \beta$).
			\STATE Update $FE$.
		\ENDWHILE
	\end{algorithmic}
	\textbf{Output:} Optimal solutions in $S_*$.
\end{algorithm}
\begin{enumerate}
	\item \textbf{Experience learning}: Before evolutionary optimization starts, a meta-learning procedure is conducted to train task-independent parameters $\bm{\gamma}^e$ for MDKL surrogates (line 1). 
	$\bm{\gamma}^e$ are trained on $N_m$ datasets $\{D_{m1}, \dots, D_{m N_m}\}$ which are collected from $N$ related tasks $\{T_1, \dots, T_N\}$. $\bm{\gamma}^e$ are experiences that represent the domain-specific features of related tasks. 
	\item \textbf{Initialize surrogates with experiences}: Evolutionary optimization starts when a target optimization task $T_*$ is given. An initial dataset $S_*$ is sampled (line 2) to adapt task-specific parameters $\bm{\gamma}^*$ on the basis of experiences $\bm{\gamma}^e$. After that, MDKL surrogates are updated (line 3).
	\item \textbf{Reproduction}: MDKL surrogates $h(\bm{\gamma}^*)$ are combined with a SAEA optimizer $Opt$ to search for optimal solution(s) $\textbf{x}^*$ on $h(\bm{\gamma}^*)$ (line 6). This is implemented by replacing the original (regression-based) surrogates in a SAEA with $h(\bm{\gamma}^*)$. 
	 \item \textbf{Update archive and surrogates}: New optimal solution(s) $\textbf{x}^*$ is evaluated on target task $T_*$ (line 7). The evaluated solutions will be added to dataset $S_*$ (line 8) which serves as an archive.
	Further surrogate adaptation is triggered, surrogates $h(\bm{\gamma}^*)$ are updated (line 9).
	\item \textbf{Stop criterion}: Once the evaluation budget has run out, the evolutionary optimization will be terminated and the optimal solutions in dataset $S_*$ will be outputted. Otherwise, the algorithm goes back to step 3.
\end{enumerate}
In the following subsections, we first present the details of our MDKL surrogate modeling method, including how to learn experiences through meta-learning and adapt it to a specific target task. Then we explain the surrogate update strategy in our experience-based SAEA framework. Finally, we discuss the usage of MDKL surrogates and the compatibility of our SAEA framework with existing SAEAs.

\subsection{Learning and Using Experiences by MDKL}
In MDKL, the domain-specific features of related tasks are used as experiences, which are represented by the task-independent parameters $\bm{\gamma}^e$ learned across related tasks. 
To make MDKL more capable of expressing complex domain-specific features, the base kernel $k(\textbf{x}^i, \textbf{x}^j |\ \bm{\theta})$ in GP is combined with a neural network $\phi(\textbf{w}, \textbf{b})$ to construct a deep kernel (see Eq.(\ref{equ: dk})).

The main novelties of our MDKL surrogate modeling method can be summarized as follows.
\begin{itemize}
	\item A simple and efficient meta-learning structure for experience learning. 
	Our method learns only one common neural network for all related tasks, such a neural network works as a component of the shared deep kernel. In comparison, \cite{Garnelo:2018:Neural, Tossou:2019:ADKL} generate separate deep kernels for each related task and train multiple neural networks to encode and decode task features, which makes their model complexity grow with the number of related tasks.
	\item The explicit task-specific adaptation for the deep kernel, which is implemented by cumulating task-specific increments on the basis of the learned experiences. 
\end{itemize}
The modeling of a MDKL model consists of two procedures: meta-learning procedure and adaptation procedure. To make a clear illustration, we introduce frameworks of two procedures and then explain them in detail.

\quad \\ \textbf{Meta-learning procedure: Learning experiences} \\ 
Our MDKL model uses the kernel in \cite{Jones:1998:EGO} as its base kernel:
\begin{equation} \label{equ: bk}
	k(\textbf{x}^i, \textbf{x}^j | \bm{\theta}, \textbf{p}) = exp(- \sum_{k=1}^d \theta_k |x_k^i - x_k^j|^{p_k}).
\end{equation} 
Therefore, the deep kernel will be:
\begin{equation} \label{equ: dk2}
	k(\textbf{x}^i, \textbf{x}^j | \bm{\gamma}) = exp(- \sum_{k=1}^d \theta_k |\phi(x_k^i) - \phi(x_k^j)|^{p_k}),
\end{equation} 
where $\bm{\gamma} = \{\textbf{w}, \textbf{b}, \bm{\theta}, \textbf{p}\}$ is a set of deep kernel parameters. 
Details about alternative base kernels are available in \cite{Rasmussen:2006:GP}. 

The aim of meta-learning procedure is to learn experiences $\bm{\gamma}^e$ from related tasks $\{T_1, \dots, T_N\}$, including neural network parameters $\textbf{w}, \textbf{b}$, and task-independent base kernel parameters $\bm{\theta}^e, \textbf{p}^e$.
The pseudo-code of meta-learning procedure is given in Algorithm \ref{alg: MDKL1}.
\begin{algorithm}[!t]
	\caption{Meta-learning($D_i, N_m, |D_m|, B, \alpha$)} \label{alg: MDKL1}
	\textbf{Input:} \\
	\hspace*{\algorithmicindent} $D_i$: Datasets collected from related tasks $T_i$, i=$\{1, \dots, N\}$; \\
	\hspace*{\algorithmicindent} $N_m$: Number of datasets $D_m$ for meta-learning; \\
	\hspace*{\algorithmicindent} $|D_m|$: Size of $D_m$. Note that $|D_m| \le |D_i|$; \\
	\hspace*{\algorithmicindent} $B$: Number of related tasks in a batch; \\
	\hspace*{\algorithmicindent} $\alpha$: Learning rate for priors. \\
	\textbf{Procedure:} 
	\begin{algorithmic}[1] 
    		\STATE Random initialize $\textbf{w}, \textbf{b}, \bm{\theta}^e, \textbf{p}^e$. 
		\STATE Set the number of update iterations U = $N_m / B$. 
		\FOR{$j=1$ to $U$}
			\STATE $\{D'_1, \dots, D'_B\} \leftarrow$ Randomly select a batch of datasets from $\{D_1, \dots, D_N\}$.
			\FORALL{$D'_i$ in the batch}
				\STATE $D_{mi} \leftarrow$ Randomly sampling($D'_i$, $|D_m|$).
				\STATE Initialize task-specific increment $\Delta \bm{\theta}^i, \Delta \textbf{p}^i$.
				\STATE Compute task-specific parameters: $\bm{\theta}^i=\bm{\theta}^e + \Delta \bm{\theta}^i$, $\textbf{p}^i = \textbf{p}^e + \Delta \textbf{p}^i$. 
				\STATE Obtain deep kernel $k(\textbf{x}^i, \textbf{x}^j | \bm{\gamma})$ based GP: $h(\bm{\gamma})$, where $\bm{\gamma}=\{\textbf{w}, \textbf{b}, \bm{\theta}^i, \textbf{p}^i\}$ (Eq.(\ref{equ: dk2})).
				\STATE Compute the loss function $L({D_{mi}}, h(\bm{\gamma}))$ (Eq.(\ref{equ: loss})).
				\STATE Update $\textbf{w}, \textbf{b}, \bm{\theta}^e, \textbf{p}^e$ using gradient descent: $\alpha \bigtriangledown L({D_{mi}}, h(\bm{\gamma}))$ (Eq.(\ref{equ: update})). 
			\ENDFOR
		\ENDFOR
	\end{algorithmic}
	\textbf{Output:} Task-independent parameters: $\bm{\gamma}^e$ = $\{\textbf{w}, \textbf{b}, \bm{\theta}^e, \textbf{p}^e\}$. 
\end{algorithm}

Ideally, the experiences $\bm{\gamma}^e$ are learned from plenty of ($N_m$) small datasets $D_m$ collected from different related tasks. However, in practice, the number of available related tasks $N$ may be much smaller than $N_m$. Hence, the meta-learning is conducted gradually over $U$ update iterations (line 2). During each update iteration, a small batch of related tasks contribute $B$ small datasets $\{D_{m1}, \dots, D_{mB}\}$ for meta-learning purpose (line 4 and line 6). Note that if $N < N_m$, a related task $T_i$ can be used multiple times in the meta-learning procedure.

For a given dataset $D_{mi}$, we denote $\bm{\theta}^i=\bm{\theta}^e+\Delta \bm{\theta}^i$ and $\textbf{p}^i=\textbf{p}^e+\Delta \textbf{p}^i$ as the task-specific kernel parameters, where $\Delta \bm{\theta}^i, \Delta \textbf{p}^i$ are the distance we need to move from the task-independent parameters to the task-specific parameters (line 8).
The loss function $L$ of MDKL is the likelihood function defined as follows \cite{Jones:1998:EGO}:
\begin{equation} \label{equ: loss}
	\frac{1}{(2\pi)^{n/2} (\sigma^2)^{n/2} |\textbf{R}|^{1/2}} exp[- \frac{(\textbf{y}-\textbf{1}\mu)^T\textbf{R}^{-1}(\textbf{y}-\textbf{1}\mu)}{2\sigma^2}],
\end{equation} 
where $|\textbf{R}|$ is the determinant of the correlation matrix $\textbf{R}$, each element in the matrix is computed through Eq.(\ref{equ: dk2}). $\textbf{y}$ is the fitness vector of $D_{mi}$. And $\mu$ and $\sigma^2$ are the mean and the variance of the prior distribution, respectively.
Experiences $\bm{\gamma}^e = \{\textbf{w}, \textbf{b}, \bm{\theta}^e, \textbf{p}^e\}$ is updated by gradient descent (line 9), take $\bm{\theta}^e$ as an example: 
\begin{equation} \label{equ: update}
	\bm{\theta}^e \leftarrow \bm{\theta}^e - \alpha \bigtriangledown_{\bm{\theta}^e} L(D_{mi}, h(\bm{\gamma})).
\end{equation} 
After $U$ iterations, $\bm{\gamma}^e$ has been trained sufficiently by $N_m$ small datasets $D_m$ and will be used in target task $T_*$ later.

\begin{figure}[!t]
	\centering
		\includegraphics[width=3.0in]{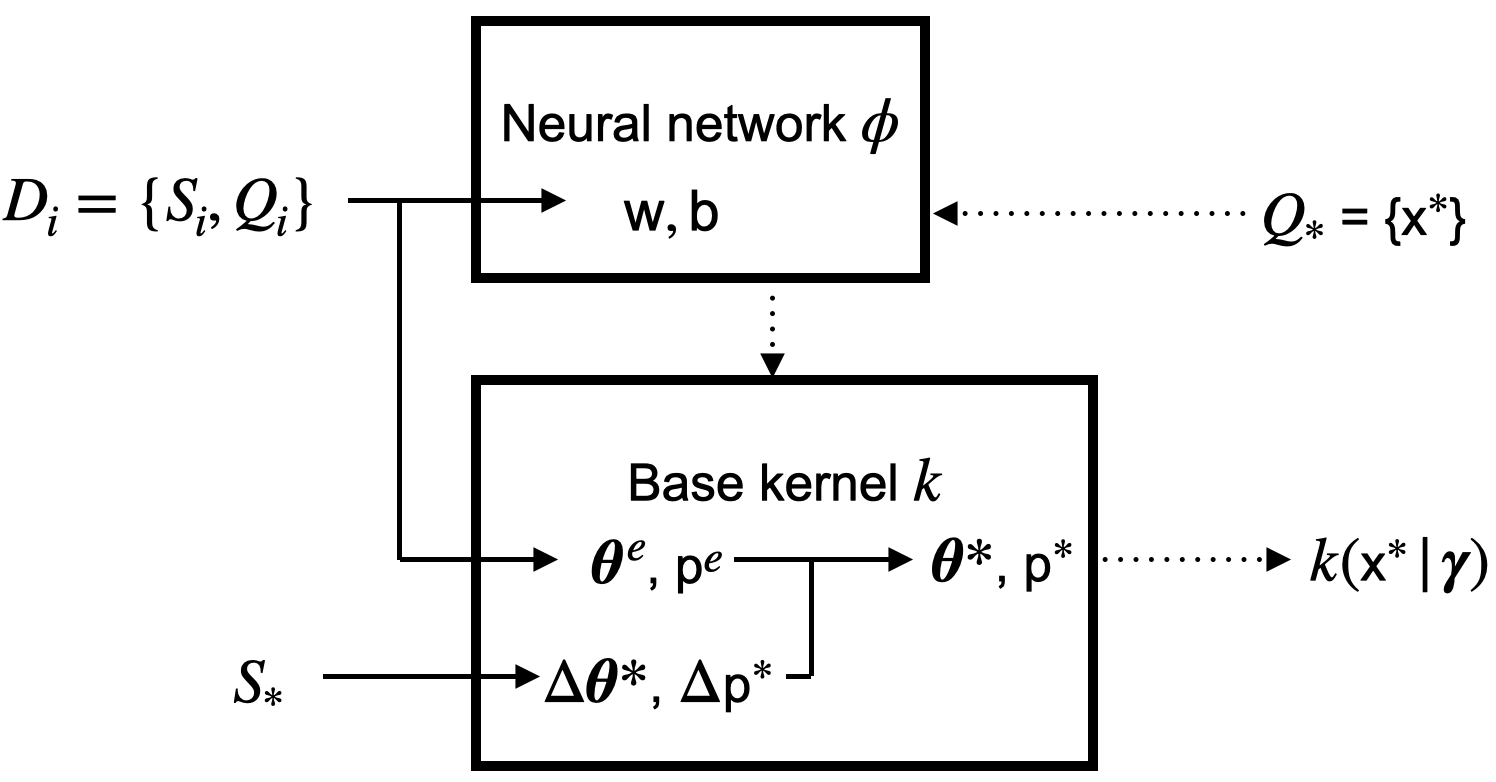}
	\caption[Diagram of our deep kernel implementation.]
	{Diagram of our deep kernel implementation. The continuous lines depict the training process, the dotted lines depict the inference process. $Q_*$ denotes query samples to be evaluated on our surrogates.
	The neural network ensures the expressive power of our deep kernel. Common features of related tasks are represented by neural network parameters $\textbf{w}, \textbf{b}$ and base kernel parameters $\bm{\theta}^e, \textbf{p}^e$, which are the learned experiences. Task-specific increments $\Delta \bm{\theta}^*$ and $\Delta \textbf{p}^*$ distinguishes a given task $T_*$ from other tasks.}
	\label{fig: diagram-structure}
\end{figure}

\quad\\ \textbf{Adaptation procedure: Using experiences} \\ 
The meta-learning of experiences $\bm{\gamma}^e$ enables MDKL to handle a family of related tasks in general. To approximate a specific task $T_*$ well, surrogate $h(\bm{\gamma}^e)$ needs to adapt task-specific increments $\Delta \bm{\theta}^*$ and $\Delta \textbf{p}^*$ in the way described in Algorithm \ref{alg: MDKL2}. 
\begin{algorithm}[!t]
	\caption{Adaptation($\bm{\gamma}^*, S_*, \beta$)}\label{alg: MDKL2}
	\textbf{Input:} \\
	\hspace*{\algorithmicindent} $\bm{\gamma}^*$: Current surrogate parameters; \\
	\hspace*{\algorithmicindent} $S_*$: A dataset sampled from target task $T_*$ (Archive). \\
	\hspace*{\algorithmicindent} $\beta$: Learning rate for adaptation. \\
	\textbf{Procedure:} 
	\begin{algorithmic}[1]
		\IF {$\bm{\gamma}^* == \bm{\gamma}^e$} 
			\STATE Initialize task-specific increments $\Delta \bm{\theta}^*, \Delta \textbf{p}^*$.  
			\STATE Compute task-specific parameters: $\bm{\theta}^*=\bm{\theta}^e + \Delta \bm{\theta}^*$, $\textbf{p}^* = \textbf{p}^e + \Delta \textbf{p}^*$. 
			\STATE Obtain deep kernel $k(\textbf{x}^i, \textbf{x}^j | \bm{\gamma}^*)$ based GP: $h(\bm{\gamma}^*)$, where $\bm{\gamma}^*=\{\textbf{w}, \textbf{b}, \bm{\theta}^*, \textbf{p}^*\}$ (Eq.(\ref{equ: dk2})).
		\ENDIF
		\STATE Compute the loss function $L(S_*, h(\bm{\gamma}^*))$ (Eq.(\ref{equ: loss})).
		\STATE Update $\Delta \bm{\theta}^*, \Delta \textbf{p}^*$ using gradient descent: $\beta \bigtriangledown$ $L(S_*, h(\bm{\gamma}^*))$ (Eq.(\ref{equ: adapt})).
	\end{algorithmic}
	\textbf{Output:} Adapted MDKL $h(\bm{\gamma}^*)$.
\end{algorithm}
There are three major differences between the meta-learning procedure and the adaptation procedure: 
\begin{itemize}
	\item First, the dataset $S_*$ for adaptations is sampled from target task $T_*$. In comparison, in the meta-learning procedure, the training datasets $\{D_{m1}, \dots, D_{mN_m}\}$ are sampled from $N$ different related tasks.
	\item Second, the parameters to be updated are task-specific increments $\Delta \bm{\theta}^*$ and $\Delta \textbf{p}^*$, not task-independent parameters $\bm{\theta}^e$ and $\textbf{p}^e$. The gradient descent in Eq.(\ref{equ: update}) is replaced by the follows (line 7).
\begin{equation} \label{equ: adapt}
	\Delta \bm{\theta}^* \leftarrow \Delta \bm{\theta}^* - \beta \bigtriangledown_{\Delta \bm{\theta}^*} L(S_*, h(\bm{\gamma}^*)).
\end{equation} 
	\item Third, task-specific increments $\Delta \bm{\theta}^*$ and $\Delta \textbf{p}^*$ are initialized only when it is the first time to adapt task-specific parameters (lines 1-5). From the perspective of tuning parameters, we can say task-independent base kernel parameters $\bm{\theta}^e$ and $\textbf{p}^e$ are the initialization points for further adaptations. In other words, the learning of a new task starts on the basis of past experiences.
\end{itemize}
A diagram of the deep kernel implemented in our MDKL model is illustrated in Fig. \ref{fig: diagram-structure}.

\subsection{Surrogate Update Strategy} 
The adaptation procedure has explained how experiences are used to adapt a MDKL surrogate with a dataset $S_*$ from the target task $T_*$. In this subsection, we describe the update strategy in our experience-based SAEA framework. 
To properly integrate experiences and data from $T_*$, our update strategy is designed to determine whether a MDKL surrogate should be adapted in the current iteration or not, ensuring an optimal update frequency of surrogates.

As illustrated in Algorithm \ref{alg: MDKL3}, the surrogate update starts when a new optimal solution(s) has been evaluated on expensive functions and an updated archive $S_*$ is available.
\begin{algorithm}[!t]
	\caption{Update($\bm{\gamma}^*, S_*, \beta$)} \label{alg: MDKL3}
	\textbf{Input:} \\
	\hspace*{\algorithmicindent} $\bm{\gamma}^*$: Current surrogate parameters; \\
	\hspace*{\algorithmicindent} $S_*$: Updated archive. \\
	\hspace*{\algorithmicindent} $\beta$: Learning rate for further adaptations. \\
	\textbf{Procedure:} 
	\begin{algorithmic}[1]
		\STATE $e_0 \gets$ MSE$(h(\bm{\gamma}^*), S_*$). 
		\STATE $h(\bm{\gamma}') \gets$ Adaptation$(\bm{\gamma}^*, S_*, \beta)$.$/^*$Temporary surrogate$^*/$
		\STATE $e_1 \gets$ MSE$(h(\bm{\gamma}'), S_*)$. 
		\IF {$e_0 > e_1$} \STATE update $\bm{\gamma}^* = \bm{\gamma}'$, obtain new $h(\bm{\gamma}^*)$.
		\ENDIF
	\end{algorithmic}
	\textbf{Output:} Optimal solutions in $S_*$.
\end{algorithm}
For a given surrogate $h(\bm{\gamma}^*)$, its mean squared error (MSE) on $S_*$ is selected as the update criterion: If the MSE after an adaptation $e_1$ (line 3) is larger than the MSE without an adaptation $e_0$ (line 1), then the surrogate will roll back to the status before the adaptation. This indicates the surrogate update has been refused and $h(\bm{\gamma}^*)$ will not be adapted in the current iteration. Otherwise, the adapted surrogate will be chosen (line 5). 
Note that no matter whether surrogate adaptations are accepted or refused, the resulting surrogates will be treated as updated surrogates, which are employed to assist the SAEA optimizer in the next iteration.

\subsection{Framework Compatibility and Surrogate Usage} \label{sec: algorithm-D}
Our experience-based SAEA framework is compatible with regression-based SAEAs since the original surrogates in these SAEAs can be replaced by our MDKL surrogates directly. 
Due to the nature of GP, when predicting the fitness of a query solution $\textbf{x}^*$, a MDKL surrogate produces a predictive Gaussian distribution $\mathcal{N}$$\sim$$(\hat{y}(\textbf{x}^*), \hat{s}^2(\textbf{x}^*))$ , the predicted mean $\hat{y}(\textbf{x}^*)$ and covariance $\hat{s}^2(\textbf{x}^*)$ are specified as \cite{Jones:1998:EGO}:
\begin{equation}
	\hat{y}(\textbf{x}^*) = \mu + \textbf{r}'\textbf{R}^{-1}(\textbf{y}-\textbf{1}\mu),
\end{equation}
\begin{equation}
	\hat{s}^2(\textbf{x}^*) = \sigma^2 (1 - \textbf{r}'\textbf{R}^{-1}\textbf{r}),
\end{equation}
where $\textbf{r}$ is a correlation vector consisting of covariances between $\textbf{x}^*$ and $S_*$, other variables are explained in Eq.(\ref{equ: loss}).

Classification-based SAEAs are not compatible with our SAEA framework. The classification surrogates in these SAEAs are employed to learn the relation between pairs of solutions, or the relation between solutions and a set of reference solutions. The class labels used for surrogate training can be fluctuating during the optimization and thus hard to be learned over related tasks. 
Similarly, in ordinal-regression-based SAEAs, the ordinal relation values to be learned are not as stable as the fitness of expensive functions. So ordinal-regression-based SAEAs are also not compatible with our SAEA framework.  
In this paper, we focus on experiences-based optimization for regression-based SAEAs, while other SAEA categories are left to be discussed in future work.

\section{Computational Studies} \label{sec: experiment} 
Our computational studies can be divided into three parts: 
\begin{itemize}
	\item Section \ref{sec: exp-MDKL} evaluates the effectiveness of learning experiences through a real-world engine modeling problem. 
	\item Sections \ref{sec: exp-DTLZ} to \ref{sec: exp-size} use the scenario of multi-objective optimization as an example to investigate the performance of our SAEA framework in depth. Empirical evidence is provided to guide the 	use of our SAEA framework.
	\item  Section \ref{sec: exp-engine} investigates the performance of our SAEA framework in breadth. The engine calibration problem we solve in the experiment covers many representative features, such as constrained optimization, single-objective optimization, and real-world applications. 
\end{itemize}
The source code is available online\footnote{https://github.com/XunzhaoYu/Experience-Based-SAEA}. 
For all meta-learning methods used in our experiments, their basic setups are listed in Table \ref{tab: setups}. 
\begin{table}[!t]
	\caption{Parameter setups for meta-learning methods.} 
	\label{tab: setups}
	\scalebox{0.95}{
	\centering
	\begin{tabular}{l l l}
	\hline Module 			& Parameter 							& Value \\
	\hline Meta-learning 		& Number of meta-learning datasets $N_m$ 	& 20000 \\
						& Number of update iterations $U$			& 2000 \\
						& Batch size $B$ 						& 10 \\
	\hline Neural network 	& Number of hidden layers 				& 2 \\
						& Number of units in each hidden layer		& 40 \\
						& Learning rates $\alpha, \beta$			& 0.001, 0.001 \\												& Activation function						& ReLU \\
	\hline
	\end{tabular}}
\end{table}
The neural network structure is suggested by \cite{Finn:2017:MAML, Patacchiola:2020:DKT}, and the learning rates are the default values that have been widely used in many meta-learning methods \cite{Harrison:2018:ALPaCA, Patacchiola:2020:DKT}.

\subsection{Effectiveness of Learning Experiences} \label{sec: exp-MDKL} 
The evaluation of the effectiveness of learning experiences aims at demonstrating that our MDKL model is able to learn experiences from related tasks and it outperforms other meta-learning models. For this reason, the experiment is designed to answer the following questions:
\begin{itemize}
	\item Given a small dataset $S_*$ from target task $T_*$, can MDKL learn experiences from related tasks and generate a model with the smallest MSE? 
	\item If yes, which components of MDKL contribute to the effectiveness of learning experiences? Meta-learning or/and deep kernel learning? If not, why not?
\end{itemize}
To answer the two questions above, we consider two experiments to evaluate the effectiveness of learning experiences: amplitude prediction for unknown periodic sinusoid functions, and fuel consumption prediction for a gasoline motor engine.
The former is a few-shot regression problem that motivates many meta-learning studies \cite{Finn:2017:MAML, Harrison:2018:ALPaCA, Tossou:2019:ADKL, Patacchiola:2020:DKT}, while the latter is a real-world regression problem \cite{Zhu:2020:ECU}. The results of the sinusoid function regression experiment are reported in the supplementary material.
In this subsection, we focus on a Brake Special Fuel Consumption (BSFC) regression task for a gasoline motor engine \cite{Zhu:2020:ECU}, where BSFC is evaluated on the gasoline engine simulation (denoted by $T_*$) provided by Ford Motor Company.

\quad \\ \textbf{Experimental setups}: \\  
The related tasks $T_i$ used in our experiment are $N=100$ gasoline engine models. These engine models have different behaviors when compared with $T_*$, but they share the basic features of gasoline engines. All related tasks and the target task have the same six decision variables.
Each related task $T_i$ provides only 60 solutions, forming a dataset $D_i$. The size of datasets used for meta-learning $|D_m|$ is set to 40.
Six tests are conducted where $|S_*|=\{$2, 3, 5, 10, 20, 40$\}$ data points are sampled from the real engine simulation $T_*$. The MSE is chosen as an indicator of modeling accuracy, which is measured using a dataset consisting of 12500 data points that are sampled uniformly from the engine decision space. 

\quad \\ \textbf{Comparison methods}: \\ 
In this experiment, three families of modeling methods are compared with our MDKL model: 
\begin{itemize}
	\item \textbf{Meta-learning methods} that are proposed for regression tasks: MAML \cite{Finn:2017:MAML}, ALPaCA \cite{Harrison:2018:ALPaCA}, and DKT \cite{Patacchiola:2020:DKT}. 
	The configurations of MAML, ALPaCA, DKT are the same as suggested in the original literature. 
	\item \textbf{Non-meta-learning method} that is widely used for regression tasks: GP model. 
	We choose GP as a baseline since it is effective and more relevant to MDKL than other non-meta-learning modeling methods.
	We set the range of base kernel parameters in the GP model as $\theta \in [10^{-5}, 10]$ and $p \in [1, 2]$. 
	\item \textbf{MDKL related methods} that are designed to investigate which components of MDKL contribute to the modeling performance: GP\_Adam, DKL, and MDKL\_NN. 
	GP\_Adam is a GP model fitted by Adam optimizer. The combination of GP\_Adam and a neural network results in a kind of DKL algorithm. MDKL\_NN is a meta-learning version of DKL, but it learns only neural network parameters through meta-learning and has no task-independent base kernel parameters. 
\end{itemize}

\quad \\ \textbf{Results and analysis}: \\  
The statistical test results of the MSE values achieved by comparison algorithms in BSFC regression experiments are summarized in Table \ref{tab: engine}. Each row lists the results obtained when the same number of fitness evaluations $|S_*|$ are used to train models. 
The results of Wilcoxon rank sum test between MDKL and other compared algorithms are listed in the last row.
\begin{table*}[!t]
	\caption[Average MSE and standard deviation (in brackets) of 30 runs on the regression of engine fuel consumption.]
	{Average MSE and standard deviation (in brackets) of 30 runs on the regression of engine fuel consumption. 
	Support data points are used to train non-meta surrogates or adapt meta-learning surrogates.
	All results are normalized since the actual engine data is unable to be disclosed. 
	The symbols '$+$', '$\approx$', '$-$' denote the win/tie/loss result of Wilcoxon rank sum test (significance level is 0.05) between MDKL and comparison modeling methods, respectively. The last row counts the total win/tie/loss results.} 
	\label{tab: engine}
	\centering
	\scalebox{0.80}{
	\begin{tabular}{l| c| c c c c| c c c}
        \hline 
	Support data	& GP \cite{Stein:1999:Kriging} 	& GP\_Adam 		&	DKL 				& 	MDKL\_NN 			&	MDKL		& DKT \cite{Patacchiola:2020:DKT}	& MAML \cite{Finn:2017:MAML}	& ALPaCA \cite{Harrison:2018:ALPaCA} \\
        points $|S_*|$ 	&					&					&					&						&				&						&						& \\
        \hline 
        	2			& 2.23e+1(3.20e+0)$+$	& 2.37e+1(6.30e+0)$+$	& 2.30e+1(5.87e+0)$+$	& 1.73e+1(6.33e+0)$\approx$	& 1.72e+1(6.34e+0)	& 1.81e+1(5.68e+0)$\approx$	& 1.87e+1(6.37e+0)$\approx$	& 1.91e+1(1.02e+1)$\approx$ \\
        	3			& 2.14e+1(3.74e+0)$+$	& 2.41e+1(1.38e+1)$+$	& 2.20e+1(3.74e+0)$+$	& 1.45e+1(7.13e+0)$\approx$	& 1.45e+0(7.01e+0)	& 1.55e+1(6.66e+0)$\approx$	& 1.80e+1(4.69e+0)$\approx$ 	& 2.13e+1(1.97e+1)$\approx$ \\
        	5			& 2.13e+1(3.27e+0)$+$	& 2.46e+1(1.00e+1)$+$	& 2.07e+1(3.95e+0)$+$	& 1.12e+1(6.65e+0)$\approx$	& 1.10e+1(6.58e+0)	& 1.21e+1(6.49e+0)$\approx$	& 1.84e+1(6.05e+0)$+$ 		& 1.99e+1(2.29e+1)$+$ \\
        	10			& 1.84e+1(1.89e+0)$+$	& 2.06e+1(1.19e+1)$+$	& 2.10e+1(5.79e+0)$+$	& 7.19e+0(4.82e+0)$\approx$	& 7.08e+0(4.77e+0)	& 7.99e+0(4.87e+0)$\approx$	& 1.70e+1(5.54e+0)$+$ 		& 1.38e+1(8.12e+0)$+$ \\	
        	20			& 1.56e+1(2.00e+0)$+$	& 2.38e+1(1.05e+1)$+$	& 1.76e+1(2.42e+0)$+$	& 5.03e+0(1.82e+0)$\approx$	& 4.86e+0(1.71e+0)	& 5.74e+0(1.91e+0)$+$		& 1.50e+1(2.59e+0)$+$ 		& 1.01e+1(5.52e+0)$+$ \\ 
        	40			& 1.28e+1(2.03e+0)$+$	& 1.48e+1(7.35e+0)$+$	& 1.67e+1(3.73e+0)$+$	& 4.13e+0(7.90e-1)$\approx$	& 4.00e+0(8.59e-1)	& 4.92e+0(1.09e+0)$+$		& 1.45e+1(1.85e+0)$+$ 		& 8.01e+0(3.35e+0)$+$ \\ 
	\hline 
	win/tie/loss 	& 6/0/0					& 6/0/0				& 6/0/0				& 0/6/0				& -/-/-			& 2/4/0					& 4/2/0					& 4/2/0 \\
	\hline
	\end{tabular}}
\end{table*}
It can be observed that MDKL and MDKL\_NN outperform other comparison modeling methods since they have achieved the smallest MSE on all tests. 
Consistent observations can also be made from the results of the sinusoid function regression experiments (presented in the supplementary material), MDKL has also achieved the best performance.

Additional Wilcoxon rank sum tests have been conducted between MDKL related algorithms to answer our second question (results are not reported in Table \ref{tab: engine}). 
The statistical test results between DKL and GP\_Adam are 1/5/0, indicating that the neural network in DKL makes some contributions to the performance of MDKL.
The statistical test results between MDKL\_NN and DKL are 6/0/0, demonstrating that the meta-learning of neural network parameters constructs a useful deep kernel and contributes to the improvement of modeling accuracy.
However, there is no significant difference between the performance of MDKL and that of MDKL\_NN, the meta-learning on base kernel parameters does not play a critical role on this engine problem. 
In comparison, the meta-learning on base kernel parameters is effective in sinusoid function regression experiments (see the supplementary material).
Besides, the statistical test results between MDKL\_NN and MAML are 4/2/0. Considering that MAML is a neural network regressor learned through meta-learning, we can conclude that GP is an essential component of our MDKL. 
In summary, all components in MDKL are necessary, and they all contribute to the effectiveness of learning experiences.

The comparison experiments on the gasoline motor engine and sinusoid functions have demonstrated the effectiveness of our MDKL modeling method in the learning of experiences. 
Given a small dataset of the target task, the model learned through MDKL method has the smallest MSE among all comparison models.
Besides, the investigation between MDKL and its variants shows that all components in MDKL have made their contributions to the effectiveness of learning experiences.
However, similar to other meta-learning studies \cite{Finn:2017:MAML, Harrison:2018:ALPaCA}, we have not defined the similarity between tasks. In other words, the boundary between related tasks and unrelated tasks has not been defined. This should be a topic of further study on meta-learning. 
Besides, the relationship between task similarity and modeling performance has not been investigated. Instead, we study the relationship between task similarity and SAEA optimization performance in Section \ref{sec: exp-out}, since our main focus is the surrogate-assisted evolutionary optimization.

\subsection{Performance on Expensive Multi-Objective Optimization} \label{sec: exp-DTLZ} 
So far we have shown the effectiveness of experience learning method. In the following subsections, we aim at demonstrating the effectiveness of our experience-based SAEA framework. 
The experiment in this subsection is designed to answer the question below:
\begin{itemize}
	\item With the experiences learned from related tasks, can our SAEA framework helps a SAEA to save 9$d$ solutions without a loss of optimization performance?
\end{itemize}
The computational study is conducted on DTLZ testing functions \cite{Deb:2005:DTLZ}. All DTLZ functions have $d=10$ decision variables and 3 objectives, as the setups that have been widely used in \cite{Pan:2018:CSEA, Song:2021:KTA2}. The details of generating DTLZ variants (related tasks) are provided in the supplementary material.

\quad \\ \textbf{Comparison algorithms}: \\ 
As explained in Section \ref{sec: algorithm-D}, our experience-based SAEA framework is compatible with regression-based SAEAs. Hence, we select MOEA/D-EGO \cite{Zhang:2010:MOEADEGO} as an example and replace its GP surrogates by our MDKL surrogates. The resulting algorithm is denoted as MOEA/D-EGO(EB). 
Note that it is not necessary to specially select a newly proposed regression-based SAEA as our example, our main objective is to save evaluations with experiences and observe if there is any damage to the optimization performance caused by the saving of evaluations. Therefore, it does not make any difference which regression-based SAEAs we choose as our example.
In addition, to demonstrate the improvement of optimization performance caused by using experiences on DTLZ functions is significant, several state-of-the-art SAEAs are also compared as baselines, including ParEGO \cite{Knowles:2006:ParEGO}, K-RVEA \cite{Chugh:2016:K-RVEA}, CSEA \cite{Pan:2018:CSEA}, OREA \cite{Yu:2019:OREA}, and KTA2 \cite{Song:2021:KTA2}. Among these SAEAs, ParEGO, K-RVEA, and KTA2 use regression-based surrogates, CSEA uses a classification-based surrogate, and OREA employs an ordinal-regression-based surrogate. 

We implemented the experience-based SAEA framework, MOEA/D-EGO, ParEGO, and OREA, while the code of K-RVEA, CSEA, and KTA2 is available on PlatEMO \cite{Tian:2017:PlatEMO}, an open source Matlab platform for evolutionary multi-objective optimization.
To make a fair comparison, all comparison algorithms share the same initial dataset $S_*$ in an independent run. We also set $\bm{\theta} \in [10^{-5}, 100]^d$ and $\textbf{p} = \textbf{2}$ for all GP surrogates as suggested in \cite{Song:2021:KTA2}, these GP surrogates are implemented through DACE \cite{Sacks:1989:DACE}. Other configurations are the same as suggested in their original literature.

\quad \\ \textbf{Experimental setups}: \\ 
The parameter setups for this multi-objective optimization experiment are listed in Table \ref{tab: DTLZ150-setups}.
\begin{table}[!t]
	\caption{Parameter setups for DTLZ optimization.} 
	\label{tab: DTLZ150-setups}
	\centering
	\scalebox{0.85}{
	\begin{tabular}{l l l}
	\hline Parameter 							& MOEA/D-EGO(EB)					& Comparisons \\
	\hline Number of related tasks $N$				& 20000 ($N_m$ in Table \ref{tab: setups})	& - \\
		 Size of datasets from related tasks $|D_i|$ 	& 20 (2$d$)							& - \\
		 Size of datasets for meta-learning $|D_m|$	& $|D_i|$								& - \\
	\hline Evaluations for initialization 				& 10 ($1d$)							& 100 (10$d$) \\
		 Evaluations for further optimization			& 50									& 50 \\
		 Total evaluations						& 60									& 150 \\
	\hline
	\end{tabular}}
\end{table}
In the meta-learning procedure, we assume that plenty of DTLZ variants are available, thus $N = N_m = 20000$, each DTLZ variant $T_i$ provides $|D_i| = |D_m| = 20$ samples for learning experiences. 
During the optimization process, an initial dataset $S_*$ is sampled using Latin-Hypervolume Sampling (LHS) method \cite{McKay:2000:LHS}, then extra evaluations are conducted until the evaluation budget has run out. 
Please note that our purpose is to use related tasks to save 9$d$ evaluations without a loss of SAEA optimization performance. Hence, the total evaluation budget for MOEA/D-EGO(EB) and comparison algorithms is different.

Since the testing functions have 3 objectives, we employ inverted generational distance plus (IGD+) \cite{Ishibuchi:2015:IGD+} as our performance indicator, where smaller IGD+ values indicate better optimization results. 5000 reference points are generated for computing IGD+ values, as suggested in \cite{Pan:2018:CSEA}.

\quad \\ \textbf{Results and analysis}: \\ 
The statistical test results are reported in Tables \ref{tab: DTLZ150}. 
\begin{table*}[!t]
	\caption[Average IGD+ values and standard deviation (in brackets) obtained from 30 runs on DTLZ functions. 50 evaluations for further optimization.]
	{Average IGD+ values and standard deviation (in brackets) obtained from 30 runs on DTLZ functions.
	MOEA/D-EGO(EB) and comparison algorithms initialize their surrogates with 10, 100 samples, respectively. Extra 50 evaluations are allowed during the optimization process.
	The symbols '$+$', '$\approx$', '$-$' denote the win/tie/loss result of Wilcoxon rank sum test (significance level is 0.05) between MOEA/D-EGO(EB) and comparison algorithms, respectively. The last row counts the total win/tie/loss results.}
	\label{tab: DTLZ150}
	\centering
	\scalebox{0.85}{
        \begin{tabular}{l| c c| c c c c c}
        \hline
        Problem	&	MOEA/D-EGO				& MOEA/D-EGO(EB)	& ParEGO 				& K-RVEA					& KTA2					& CSEA					& OREA	\\
        \hline
        DTLZ1		& 1.07e+2(2.05e+1)$+$		& 9.70e+1(1.87e+1)		& 7.82e+1(1.54e+1)$-$		& 1.18e+2(2.45e+1)$+$		& 1.01e+2(2.38e+1)$\approx$	& 1.10e+2(2.50e+1)$+$ 		& 1.02e+2(1.97e+1)$\approx$ \\
	DTLZ2		& 2.99e-1(7.01e-2)$+$		& 1.43e-1(2.29e-2)		& 3.17e-1(4.12e-2)$+$		& 2.69e-1(5.97e-2)$+$ 		& 2.14e-1(3.84e-2)$+$  		& 2.98e-1(5.25e-2)$+$  		& 1.76e-1(4.69e-2)$+$ \\	
	DTLZ3		& 3.15e+2	(6.04e+1)$+$		& 1.97e+2	(1.64e+1)		& 2.30e+2	(5.99e+1)$\approx$	& 3.24e+2	(5.90e+1)$+$ 		& 2.67e+2	(6.70e+1)$+$  		& 2.82e+2(6.97e+1)$+$  		& 2.72e+2(6.88e+1)$+$ \\   
    	DTLZ4		& 5.04e-1(8.25e-2)$\approx$	& 4.44e-1(1.35e-1)		& 5.44e-1(7.58e-2)$+$		& 4.57e-1(1.14e-1)$\approx$	& 4.51e-1(9.54e-2)$\approx$ 	& 4.75e-1(1.09e-1)$\approx$	& 3.18e-1(1.54e-1)$-$ \\ 
    	DTLZ5		& 2.39e-1(7.17e-2)$+$		& 1.13e-1(2.24e-2) 		& 2.58e-1(3.68e-2)$+$		& 1.92e-1 (5.97e-2)$+$		& 1.44e-1(4.60e-2)$+$  		& 2.14e-1(4.05e-2)$+$ 		& 7.84e-2(2.42e-2)$-$ \\ 
    	DTLZ6		& 1.29e+0(4.74e-1)$\approx$	& 1.11e+0(5.71e-1)		& 1.67e+0(6.77e-1)$+$		& 4.62e+0(6.42e-1)$+$		& 3.37e+0(6.71e-1)$+$ 		& 6.26e+0(3.40e-1)$+$  		& 4.60e+0(1.19e+0)$+$ \\
    	DTLZ7		& 3.31e-1(3.11e-1)$-$ 		& 2.47e+0(1.89e+0) 		& 3.66e-1(1.31e-1)$-$ 		& 1.74e-1(3.57e-2)$-$		& 4.34e-1(2.20e-1)$-$ 		& 4.17e+0(1.13e+0)$+$ 		& 2.14e+0(1.15e+0)$\approx$ \\  
	\hline  			
	win/tie/loss 	& 4/2/1					& -/-/-				& 4/1/2					& 5/1/1					& 4/2/1					& 6/1/0					& 3/2/2 \\ 
	\hline  		
	\end{tabular}}
\end{table*}
	It can be seen from Table \ref{tab: DTLZ150} that, although 90 fewer evaluations are used in surrogate initialization, MOEA/D-EGO(EB) can still achieve competitive or even smaller IGD+ values than MOEA/D-EGO on all DTLZ functions except for DTLZ7. 
	Fig. \ref{supp-fig: DTLZ150} (in the supplementary material) also shows that the IGD+ values obtained by MOEA/D-EGO(EB) drop rapidly, especially during the first few evaluations, implying the experience learned from DTLZ variants are effective. 
	Therefore, in most situations, our experience-based SAEA framework is able to assist MOEA/D-EGO in reaching competitive or even better optimization results, with the number of evaluations used for surrogate initialization reduced from 10$d$ to only 1$d$. 
	
	MOEA/D-EGO(EB) is less effective on DTLZ7 than on other DTLZ functions, which might be attributed to the discontinuity of the Pareto Front on DTLZ7. 
	Note that MOEA/D-EGO(EB) learns experience from small datasets such as $D_m$ and $S_*$. The solutions in these small datasets are sampled randomly, hence, the probability of having optimal solutions being sampled is small. However, it is difficult to learn the discontinuity of the Pareto Front from the sampled non-optimal solutions. As a result, the knowledge of 'there are four discrete optimal regions' cannot be learned from such small datasets ($|D_m|=20$) collected from related tasks.
	
The experience learned from related tasks makes MOEA/D-EGO more competitive when compared to other SAEAs. 
	The use of MDKL surrogates results in significantly smaller IGD+ values on DTLZ1, DTLZ2, DTLZ3, and DTLZ5 than before. As a result, MOEA/D-EGO(EB) achieves the smallest IGD+ values on DTLZ2 and DTLZ3, and its optimization results on DTLZ1 and DTLZ5 are much closer to the best optimization results (e.g. results obtained by ParEGO and OREA) than MOEA/D-EGO.
	Although MOEA/D-EGO(EB) does not achieve the smallest IGD+ values on all DTLZ functions, it should be noted that MOEA/D-EGO(EB) is still the best algorithm among comparison SAEAs due to its overall performance. 
	Table \ref{tab: DTLZ150} shows that no comparison SAEA outperforms MOEA/D-EGO(EB) on three DTLZ functions, but MOEA/D-EGO(EB) outperforms all comparison SAEAs on at least three DTLZ functions.
	 Furthermore, the IGD+ values of MOEA/D-EGO(EB) are achieved with an evaluation budget of 60, while the IGD+ values of other SAEAs are reached with a cost of 150 evaluations (see Table \ref{tab: DTLZ150-setups}).
	
\quad \\ \textbf{More performance comparison experiments}: \\ 
	We also compared our MOEA/D-EGO(EB) with the best two comparison algorithms in Table \ref{tab: DTLZ150}, ParEGO and OREA, using the same evaluation budget. The results reported in Table S\ref{supp-tab: DTLZadditional} (see the supplementary material) show that our experience-based SAEA framework is more effective than comparison algorithms when the same evaluation budget is used.
Besides, the performance of our SAEA framework in the context of extremely expensive optimization has been investigated in the supplementary material (see Table S\ref{supp-tab: DTLZ90}).

The question raised at the beginning of this subsection can be answered by the results discussed so far. 
	Due to the integration of experiences learned from related tasks (DTLZ variants), although the evaluation cost of surrogates initialization has been reduced from 10$d$ to 1$d$, our experience-based SAEA framework is still capable of assisting regression-based SAEAs to achieve competitive or even better optimization results in most situations.

\subsection{Influence of Task Similarity} \label{sec: exp-out} 
In real-world applications, it might be unrealistic to assume that some related tasks are very similar to the target task. A more common situation is that all related tasks have limited similarity to the target task. 
To investigate the relationship between task similarity and SAEA optimization performance, we also tested the performance in an `out-of-range' situation, where the original DTLZ is excluded from the range of DTLZ variants during the MDKL meta-learning procedure. As a result, only the DTLZ variants that are quite different from the original DTLZ function can be used to learn experiences. The `out-of-range' situation eliminates the probability that MDKL surrogates benefit greatly from the DTLZ variants that are very similar to the original DTLZ function. 
Detailed definitions of the related tasks used in the `out-of-range' situation are given in the supplementary material. Apart from the related tasks used, the remaining experimental setups are the same as the setups described in Section \ref{sec: exp-DTLZ}. For the sake of convenience, we denote the situation we tested in Section \ref{sec: exp-DTLZ} as `in-range' below. 

The statistical test results reported in Table \ref{tab: DTLZ60-out} 
\begin{table}[!t]
	\caption[Average IGD+ values and standard deviation (in brackets) obtained from 30 runs on DTLZ functions. Comparison between in-range and out-of-range.]
	{Average IGD+ values and standard deviation (in brackets) obtained from 30 runs on DTLZ functions.
	Both MOEA/D-EGO(EB)s initialize their surrogates with 10 samples, extra 50 evaluations are allowed in the further optimization.
	The symbols '$+$', '$\approx$', '$-$' denote the win/tie/loss result of Wilcoxon rank sum test (significance level is 0.05) between the `out-of-range' and the `in-range' situations, respectively.
	The last two rows count the statistical test results between MOEA/D-EGO(EB)s and other compared algorithms.}
	\label{tab: DTLZ60-out}
	\centering
	\scalebox{0.95}{
        \begin{tabular}{l c c}
        \hline
        MOEA/D-EGO(EB)s		&	In-range				& Out-of-range	\\
        \hline
        DTLZ1				& 9.70e+1(1.87e+1)$\approx$ 	& 9.11e+1(1.53e+1) \\
        DTLZ2				& 1.43e-1(2.29e-2)$\approx$ 	& 1.41e-1(1.75e-2) \\
	DTLZ3				& 1.97e+2	(1.64e+1)$\approx$ 	& 1.98e+1(1.51e+1) \\
	DTLZ4				& 4.44e-1(1.35e-1)$\approx$ 	& 4.96e-1(8.63e-2) \\  
	DTLZ5				& 1.13e-1(2.24e-2)$\approx$	& 1.03e-1(2.39e-2) \\
	DTLZ6				& 1.11e+0(5.71e-1)$\approx$ 	& 1.17e+0(6.88e-1) \\	 	 
	DTLZ7				& 2.47e+0(1.89e+0)$\approx$ 	& 2.86e+0(1.87e+0) \\			  
	\hline  			
	win/tie/loss 			& 0/7/0					& -/-/- \\
	\hline
	vs MOEA/D-EGO		& 4/2/1					& 4/2/1 \\
	vs 6 Comparisons		& 26/9/7					& 27/7/8 \\ 
	\hline  		
	\end{tabular}}
\end{table}
show that the `out-of-range' situation achieves competitive IGD+ values to the `in-range' situation on all 7 test instances. This suggests that related tasks that are very similar to the target task have a limited impact on the optimization performance of our experience-based SAEA framework. Useful experiences can be learned from related tasks that are not very similar to the target task.
Crucially, when comparing the performance of the `out-of-range' situation and that of MOEA/D-EGO, we can still observe competitive or improved optimization results on 6 DTLZ functions (see Table \ref{tab: DTLZ60-out}, the row titled by 'vs MOEA/D-EGO', or Fig. \ref{supp-fig: DTLZ150} in the supplementary material). 
Moreover, it can be seen from the last row of Table \ref{tab: DTLZ60-out} that the `out-of-range' situation achieves better/competitive/worse IGD+ values than all compared SAEAs on 27/7/8 test instances. In comparison, the corresponding statistical test results for the `in-range' situation are 26/9/7. 
The difference between these statistical test results is not significant. 

A study on the `out-of-range' situation in the context of extremely expensive multi-objective optimization is presented in Section \ref{supp-sec: exp-out90} of the supplementary material. Consistent results can be observed from Table S\ref{supp-tab: DTLZ40-out} and Fig. \ref{supp-fig: DTLZ90}.
	
Consequently, related tasks that are very similar to the target task are not essential to the optimization performance of our experience-based SAEA framework. 
In the `out-of-range' situation, our MOEA/D-EGO(EB) can still achieve competitive or better optimization results than MOEA/D-EGO while using only 1$d$ samples for surrogate initialization.

\subsection{Influence of the Size of Datasets Used in Meta-Learning}  \label{sec: exp-size} 
We also investigated the performance of our experience-based SAEA framework when different sizes of datasets $|D_m|$ are used in the meta-learning procedure. The experimental setups are the same as the setups of MOEA/D-EGO(EB) in Section \ref{sec: exp-DTLZ} except for $|D_m|$.
\begin{table}[!t]
	\caption[Average IGD+ values and standard deviation (in brackets) obtained from 30 runs on DTLZ functions. Different sizes for experience learning.]
	{Average IGD+ values and standard deviation (in brackets) obtained from 30 runs on DTLZ functions.
	10 samples are used for initialization and extra 50 evaluations are allowed in the further optimization. $|D_m|$ is the size of the dataset collected from each related task. 
	The symbols '$+$', '$\approx$', '$-$' denote the win/tie/loss result of Wilcoxon rank sum test (significance level is 0.05) between $|D_m|$=60 and $|D_m|$=20, respectively. The last row counts the total win/tie/loss results.}
	\label{tab: DTLZ-D_m}
	\centering
	\scalebox{0.75}{
	\begin{tabular}{l c c c c}
        \hline
	Problem		&	\multicolumn{2}{c}{In-range}					&	 \multicolumn{2}{c}{Out-of-range}			\\
				&	$|D_m|$=20			&	$|D_m|$=60		&	$|D_m|$=20			&	$|D_m|$=60	\\	
        \hline
        DTLZ1		& 9.70e+1(1.87e+1)$\approx$ 	& 9.77e+1(1.73e+1)		& 9.11e+1(1.53e+1)$\approx$ 	& 9.93e+1(1.87e+1) \\
        DTLZ2		& 1.43e-1(2.29e-2)$+$ 		& 1.24e-1(2.11e-2)		& 1.41e-1(1.75e-2)$+$		&  1.29e-1(2.36e-2) \\
	DTLZ3		& 1.97e+2	(1.64e+1)$\approx$ 	& 1.98e+2	(2.21e+1)		& 1.98e+1(1.51e+1)$\approx$	& 1.93e+2(1.19e+1) \\
	DTLZ4		& 4.44e-1(1.35e-1)$\approx$ 	& 5.17e-1(5.68e-2)		& 4.96e-1(8.63e-2)$\approx$	& 5.17e-1(5.38e-2) \\  
	DTLZ5		& 1.13e-1(2.24e-2)$+$		& 9.96e-2(2.18e-2)		& 1.03e-1(2.39e-2)$\approx$	& 1.05e-1(2.73e-2) \\
	DTLZ6		& 1.11e+0(5.71e-1)$\approx$ 	& 1.04e+0(6.06e-1)		& 1.17e+0(6.88e-1)$\approx$	& 1.22e+0(6.41e-1) \\	 	 	
	DTLZ7		& 2.47e+0(1.89e+0)$+$ 		& 7.49e-1(2.61e-1)		& 2.86e+0(1.87e+0)$+$		&  6.96e-1(2.41e-1) \\			
        \hline
        win/tie/loss	& 3/4/0					&	-/-/-				& 2/5/0					&	-/-/-		\\
        \hline
	\end{tabular}}
\end{table}

	It is evident from Table \ref{tab: DTLZ-D_m} that when each DTLZ variant provides $|D_m|$=60 samples for the meta-learning of MDKL surrogates, the performance of both MOEA/D-EGO(EB)s are improved on 2 or 3 DTLZ functions.
	Particularly, a significant improvement can be observed from the optimization results of DTLZ7. As we discussed in Section \ref{sec: exp-DTLZ}, the poor performance of our experience-based optimization on DTLZ7 is caused by the small $|D_m|$. Optimal solutions have few chances to be included in a small $D_m$, which makes $D_m$ fails to provide experiences about the discontinuity of optimal regions. In comparison, the experience of 'optimal regions' can be learned from large datasets $D_m$ and thus the optimization results are improved significantly. 

	In conclusion, for our experience-based SAEA framework, a large $|D_m|$ for meta-learning procedure indicates more useful experiences can be learned from related tasks, which further improves the performance of experience-based optimization. Therefore, when applying our experience-based SAEA framework to real-world optimization problems, it is preferable to collect more data from related tasks for experience learning.

\subsection{Performance on Expensive Constrained Optimization (Engine Calibration Problem)} \label{sec: exp-engine} 
The experiments on multi-objective benchmark testing functions have investigated the performance of our SAEA framework in depth. In this subsection, we use our proposed framework to solve a real-world gasoline motor engine calibration problem, which is an expensive constrained single-objective optimization problem. This experiment covers the optimization scenarios of single-objective optimization, constrained optimization, and real-world applications. 
It serves as an example to demonstrate the generality and broad applicability of our proposed framework.

The calibration problem has 6 adjustable engine parameters, namely the throttle angle, waste gate orifice, ignition timing, valve timings, state of injection, and air-fuel-ratio. 
The calibration aims at minimizing the BSFC while satisfying 4 constraints in terms of temperature, pressure, CA50, and load simultaneously \cite{Zhu:2020:ECU}. 

\quad \\ \textbf{Comparison algorithms}: \\
Since the comparison algorithms in the DTLZ optimization experiments are not designed for handling constrained single-objective optimization, our comparison is conducted with two state-of-the-art constrained optimization algorithms used in industry \cite{Zhu:2020:ECU}: a variant of EGO designed to handle constrained optimization problems (denoted by cons\_EGO), and a GA customized for this calibration problem (denoted by adaptiveGA). The settings of comparison algorithms are the same as suggested in \cite{Zhu:2020:ECU}.  
In this experiment, we apply our SAEA framework to cons\_EGO and investigate its optimization performance. The GP surrogates in cons\_EGO are replaced by our MDKL surrogates to conduct the comparison, and the resulting algorithm is denoted as cons\_EGO(EB). 

\quad \\ \textbf{Experimental setups}: \\
The setup of related tasks ($N, D_i$) is the same as described in Section \ref{sec: exp-MDKL}.
In the meta-learning procedure, both the support set and the query set contain 6 data points, thus $|D_m| = 12$.
The total evaluation budget for all algorithms is set to 60. 
For adaptiveGA, all evaluations are used in the optimization process as it is not a SAEA. 
For cons\_EGO, 40 samples are used to initialize surrogates and 20 extra evaluations are used in the optimization process. 
For cons\_EGO(EB), only 6 samples are used to initialize MDKL surrogates, and the remaining evaluations are used for further optimization. 

\begin{figure}[!t]
	\centerline{
	\subfloat[]{\includegraphics[width=2.0in]{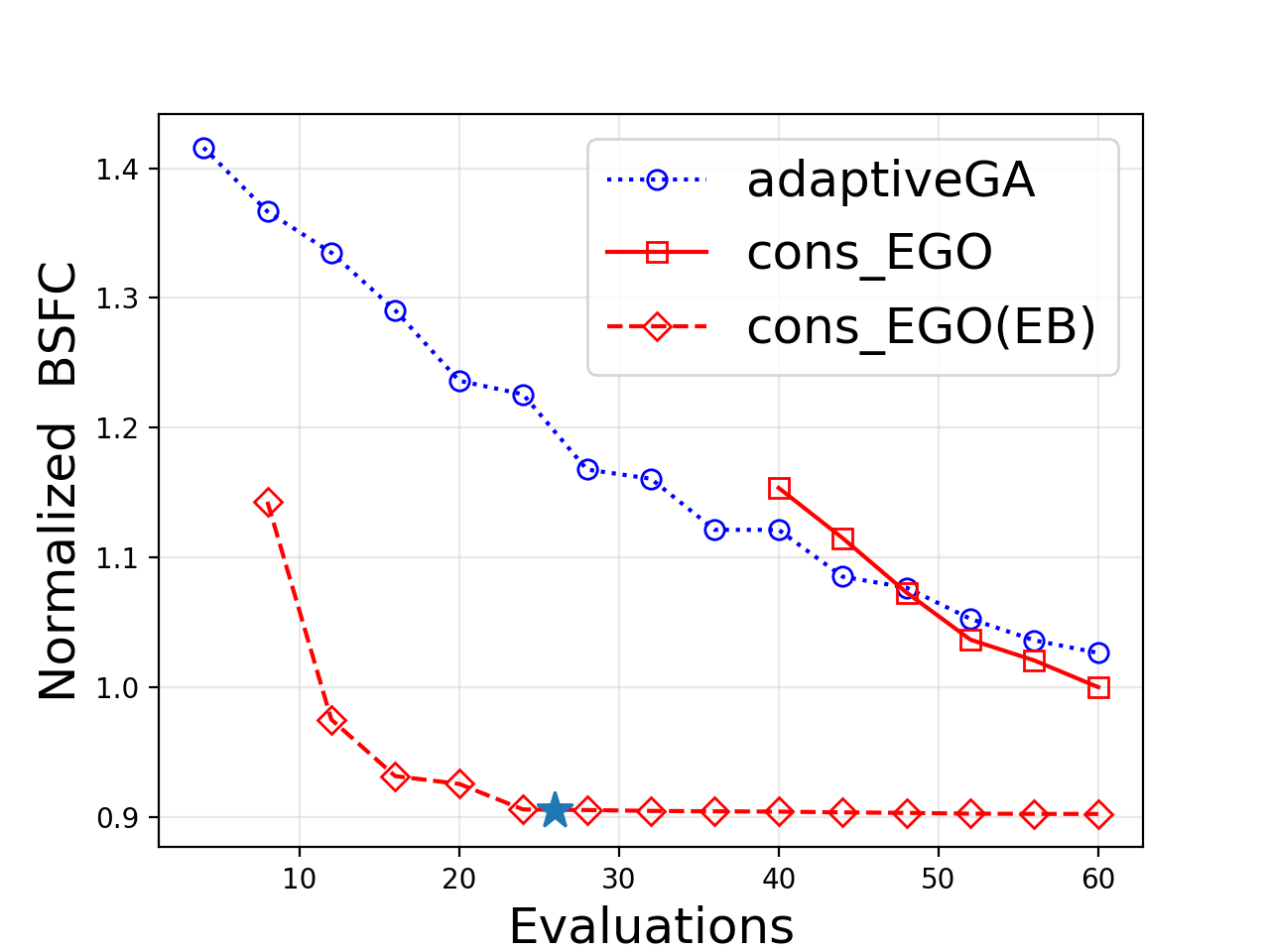}%
	\label{fig: engine_BSFC1}} \hspace*{-1.5em}
	\subfloat[]{\includegraphics[width=2.0in]{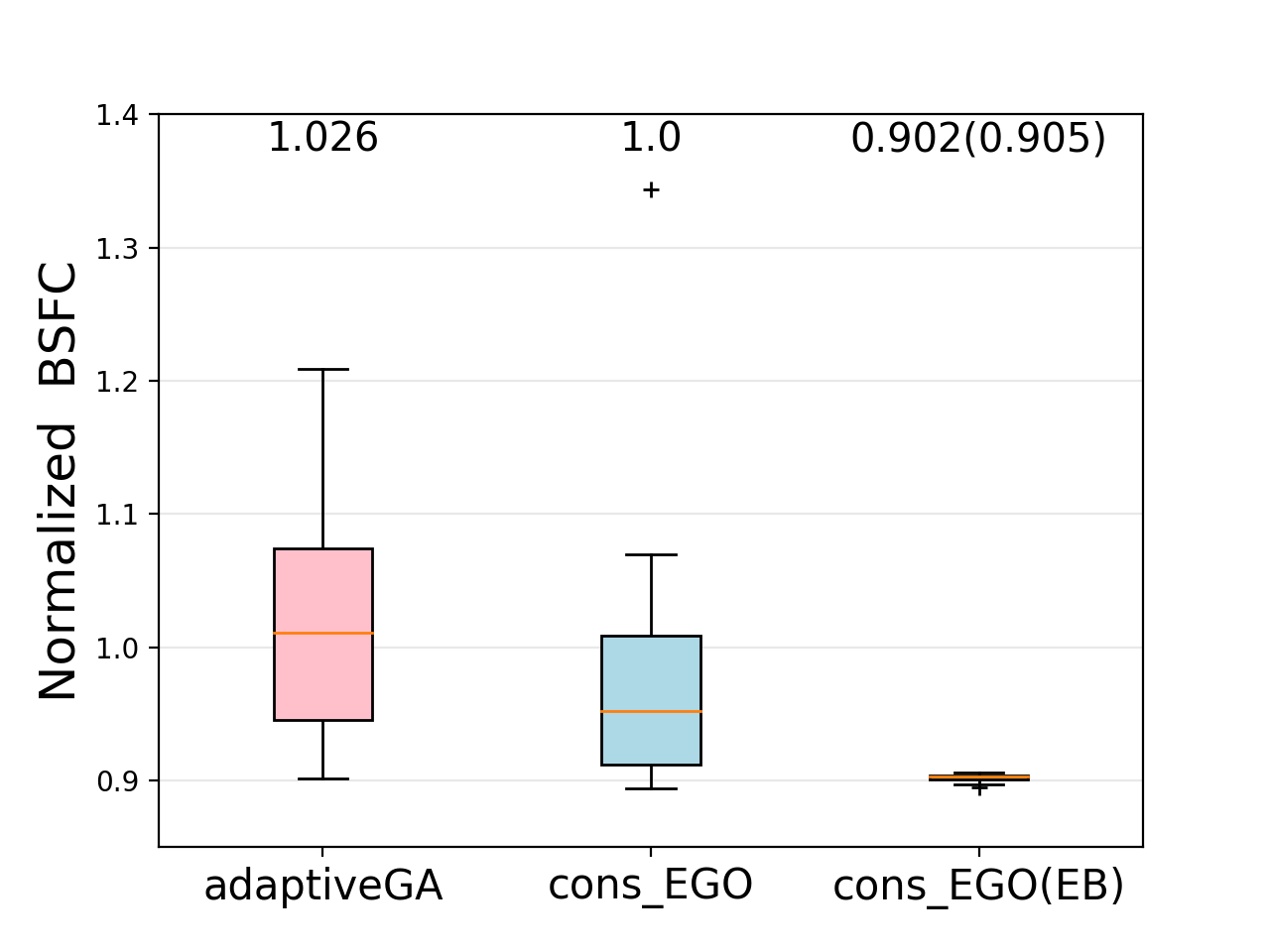}%
	\label{fig: engine_BSFC2}}
	}
	\hfil
	\centerline{
	\subfloat[]{\includegraphics[width=2.0in]{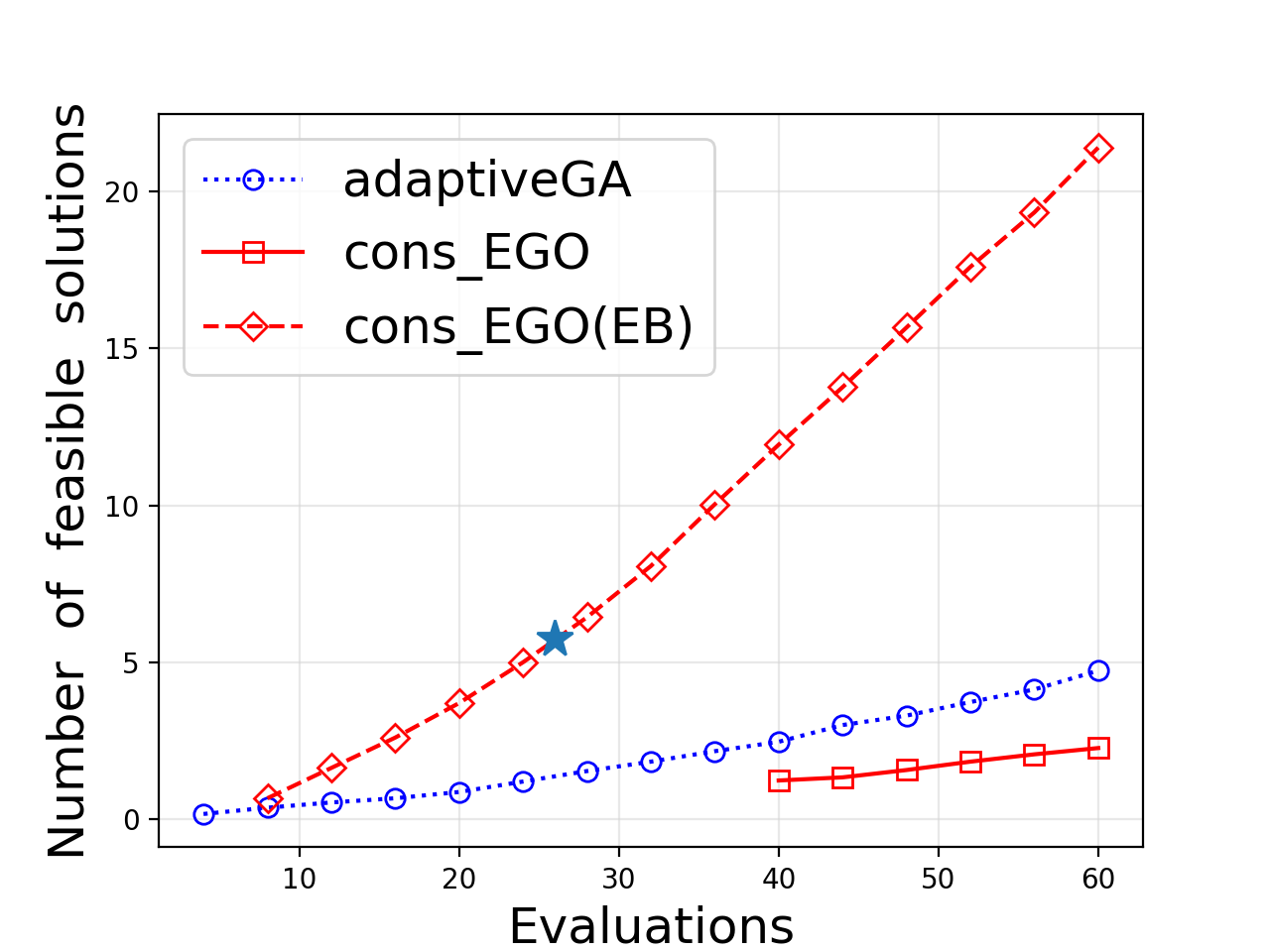}%
	\label{fig: engine_feasible1}} \hspace*{-1.5em}
	\subfloat[]{\includegraphics[width=2.0in]{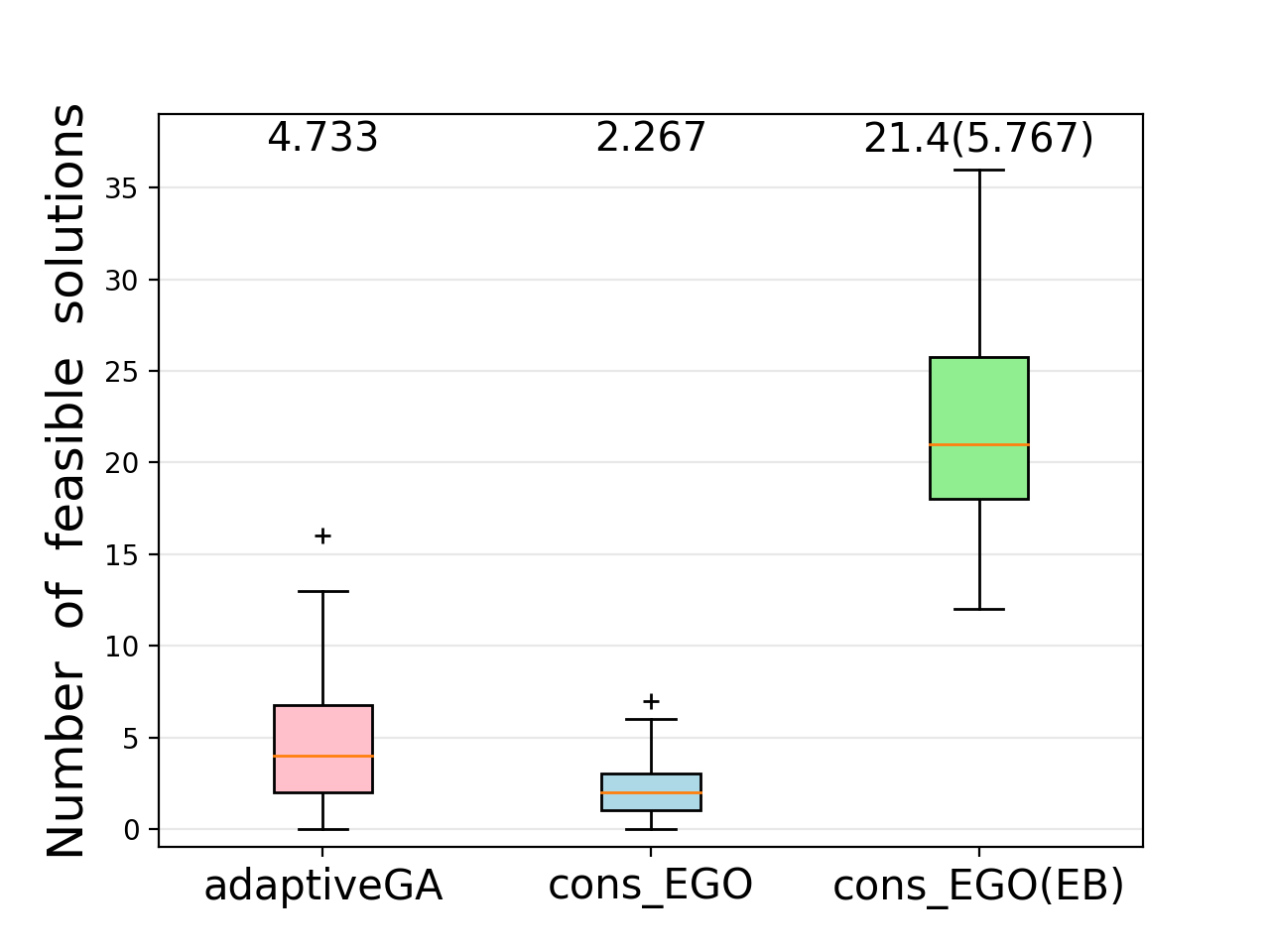}%
	\label{fig: engine_feasible2}}
	}
	\caption[Results of 30 runs on the engine calibration problem, all BSFC values are normalized.]
	{Results of 30 runs on the engine calibration problem, all BSFC values are normalized. 
	The evaluation budget is set to 60, including 40, 6 samples used to initialize surrogates for cons\_EGO and its experience-based variant, respectively. 
	Figs. (a) and (c) show how BSFC and the number of feasible solutions vary with the number of evaluations, respectively. The star markers highlight the results achieved when 20 evaluations are used in the optimization process. 
	Figs. (b) and (d) illustrate the statistical results of BSFC and the number of feasible solutions when the evaluation budget has run out. Mean values are shown on the top, the results in brackets are achieved at the star markers of Figs. (a) and (c).} 
	\label{fig: Engine}
\end{figure}

\quad \\ \textbf{Optimization results and analysis}: \\
The average normalized BSFC results and the average number of feasible solutions found over the number of evaluations used are plotted in Figs. \ref{fig: engine_BSFC1} and \ref{fig: engine_feasible1}, respectively. The statistical results of three comparison algorithms are illustrated in Figs. \ref{fig: engine_BSFC2} and \ref{fig: engine_feasible2}.
	From Fig. \ref{fig: engine_BSFC1}, it can be observed that the minimal BSFC obtained by cons\_EGO(EB) decreases drastically in the first few evaluations, implying that the experiences learned from related tasks are effective. In comparison, the minimal BSFC obtained by adaptiveGA and cons\_EGO drops in a relatively slower rate, even though cons\_EGO has used 34 more samples to initialize its surrogates. 
The star marker denotes the point at which cons\_EGO(EB) has evaluated 20 samples after surrogate initialization. It is worth noting that when 20 samples have been evaluated in the optimization, cons\_EGO(EB) achieves a smaller BSFC value than cons\_EGO. 
After the star marker, the decrease of BSFC becomes slower as cons\_EGO(EB) has reached the optimal region. Therefore, further improvement in the normalized BSFC value is not significant and thus hard to observe. 
	The advantages of our experience-based SAEA framework can also be observed in constraint handling. In Figs. \ref{fig: engine_feasible1} and \ref{fig: engine_feasible2}, cons\_EGO(EB) finds more feasible solutions than two comparison algorithms.
	These results indicate that our SAEA framework improves the performance of cons\_EGO on both objective function and constraint functions. Meanwhile, only 1$d$ evaluations are used to initialize surrogates.
	
\quad \\ \textbf{Discussion on runtime}: \\
It should be noted that real engine performance evaluations on engine facilities are very costly in terms of both time and financial budget \cite{Yu:2022:ECS}. 
Since a single real engine performance evaluation can cost several hours \cite{Ma:2013:COE, Yu:2022:ECS}, the time cost of the meta-learning procedure is negligible as it takes only a few minutes. 
Savings from reduced real engine performance evaluations on engine facilities and the reduced development cycle due to our SAEA framework could amount to millions of dollars \cite{Yu:2022:ECS}. 
Our SAEA framework is an effective and efficient method to solve this real-world calibration problem.

\section{Conclusion and further work} \label{sec: conclusion} 
Experienced human engineers are good at tackling new and unseen optimization problems in comparison to novices. There has been an on-going effort in evolutionary optimization trying to capture and then use such experiences \cite{Liu:2017:EBO, Tang:2021:FSP, Tan:2021:ETO}.
In this paper, we present an experience-based SAEA framework to solve expensive optimization problems. 
To learn experiences from related tasks, a novel meta-learning modeling method, namely MDKL, has been developed. Our MDKL model learns the domain-specific features of a set of related tasks from plenty of small datasets. 
The learned experiences are integrated with very limited examples collected from the target optimization problem, which improves the modeling efficiency and approximation accuracy. The effectiveness of learning experiences has been demonstrated by comparing our MDKL modeling method with other meta-learning and non-meta-learning modeling methods.
Our experience-based SAEA framework uses MDKL models as surrogates and an MSE-based update criterion is proposed to manage MDKL surrogates during the evolutionary optimization. Our SAEA framework is applicable to any regression-based SAEAs by replacing their original surrogates with our MDKL surrogates.
Our computational studies have demonstrated the effectiveness of our SAEA framework on multi-objective optimization and constrained single-objective optimization (a real-world engine application). On most testing functions, better or comparable optimization results are achieved when only 1$d$ samples are used to initialize our surrogates. 
A detailed summary of our experiments is presented in the supplementary material.

There are still many open research questions remaining. For example, the precise definition of experience is still unclear. Different work so far seems to have capture different aspects of experience. A more precise and comprehensive definition of optimization experience is needed. It is also a challenging task to represent such experience formally. Regarding to the SAEA work in this paper, there are two specific future research directions. 
	First, we do not have a mathematical definition of related tasks. 
Although our computational studies have demonstrated that the experiences learned from a set of related tasks that differ from the target task are beneficial to the optimization, we cannot guarantee the optimization performance if we further decrease the similarity between the related tasks and the target task. 
It is very interesting to study similarity measures between tasks in the context of experience-based SAEA framework. 
In fact, this is also a general research topic that is relevant to other experience-based optimization methods, such as transfer optimization \cite{Ruan:2019:SSCI, Ruan:2020:CEC}.
	Second, the proposed framework is currently for regression-based SAEAs only. As we can see from the DTLZ optimization experiments, classification-based SAEAs and ordinal-regression-based SAEAs could be more effective sometimes. These SAEAs learn surrogates from user-assigned class labels or ordinal relationships, instead of original fitness values. It is an interesting future work to do meta-learning from user-assigned values.

\ifCLASSOPTIONcaptionsoff
  \newpage
\fi

\bibliographystyle{IEEEtran}
\bibliography{ExperienceBasedEA.bib}

\begin{thebibliography}{10}
\providecommand{\url}[1]{#1}
\csname url@samestyle\endcsname
\providecommand{\newblock}{\relax}
\providecommand{\bibinfo}[2]{#2}
\providecommand{\BIBentrySTDinterwordspacing}{\spaceskip=0pt\relax}
\providecommand{\BIBentryALTinterwordstretchfactor}{4}
\providecommand{\BIBentryALTinterwordspacing}{\spaceskip=\fontdimen2\font plus
\BIBentryALTinterwordstretchfactor\fontdimen3\font minus
  \fontdimen4\font\relax}
\providecommand{\BIBforeignlanguage}[2]{{%
\expandafter\ifx\csname l@#1\endcsname\relax
\typeout{** WARNING: IEEEtran.bst: No hyphenation pattern has been}%
\typeout{** loaded for the language `#1'. Using the pattern for}%
\typeout{** the default language instead.}%
\else
\language=\csname l@#1\endcsname
\fi
#2}}
\providecommand{\BIBdecl}{\relax}
\BIBdecl

\bibitem{Liu:2017:EBO}
S.~Liu, K.~Tang, and X.~Yao, ``Experience-based optimization: A coevolutionary
  approach,'' \emph{arXiv preprint arXiv:1703.09865}, 2017.

\bibitem{Tang:2021:FSP}
K.~Tang, S.~Liu, P.~Yang, and X.~Yao, ``Few-shots parallel algorithm portfolio
  construction via co-evolution,'' \emph{IEEE Transactions on Evolutionary
  Computation}, vol.~25, no.~3, pp. 595--607, 2021.

\bibitem{Tan:2021:ETO}
K.~C. Tan, L.~Feng, and M.~Jiang, ``Evolutionary transfer optimization-a new
  frontier in evolutionary computation research,'' \emph{IEEE Computational
  Intelligence Magazine}, vol.~16, no.~1, pp. 22--33, 2021.

\bibitem{Jiang:2020:AFD}
M.~Jiang, Z.~Wang, L.~Qiu, S.~Guo, X.~Gao, and K.~C. Tan, ``A fast dynamic
  evolutionary multiobjective algorithm via manifold transfer learning,''
  \emph{IEEE Transactions on Cybernetics}, vol.~51, no.~7, pp. 3417--3428,
  2020.

\bibitem{Jiang:2020:IBT}
M.~Jiang, Z.~Wang, S.~Guo, X.~Gao, and K.~C. Tan, ``Individual-based transfer
  learning for dynamic multiobjective optimization,'' \emph{IEEE Transactions
  on Cybernetics}, vol.~51, no.~10, pp. 4968--4981, 2020.

\bibitem{Wei:2021:MTR}
T.~Wei, S.~Wang, J.~Zhong, D.~Liu, and J.~Zhang, ``A review on evolutionary
  multi-task optimization: Trends and challenges,'' \emph{IEEE Transactions on
  Evolutionary Computation}, vol.~26, no.~5, pp. 941--960, 2021.

\bibitem{Bali:2019:MFEAII}
K.~K. Bali, Y.-S. Ong, A.~Gupta, and P.~S. Tan, ``Multifactorial evolutionary
  algorithm with online transfer parameter estimation: {MFEA-II},'' \emph{IEEE
  Transactions on Evolutionary Computation}, vol.~24, no.~1, pp. 69--83, 2019.

\bibitem{Xue:2020:ATE}
X.~Xue, K.~Zhang, K.~C. Tan, L.~Feng, J.~Wang, G.~Chen, X.~Zhao, L.~Zhang, and
  J.~Yao, ``Affine transformation-enhanced multifactorial optimization for
  heterogeneous problems,'' \emph{IEEE Transactions on Cybernetics}, pp. 1--15,
  2020.

\bibitem{Ruan:2019:SSCI}
G.~Ruan, L.~L. Minku, S.~Menzel, B.~Sendhoff, and X.~Yao, ``When and how to
  transfer knowledge in dynamic multi-objective optimization,'' in
  \emph{Proceedings of the 2019 IEEE Symposium Series on Computational
  Intelligence (SSCI'19)}, 2019, pp. 2034--2041.

\bibitem{Ruan:2020:CEC}
------, ``Computational study on effectiveness of knowledge transfer in dynamic
  multi-objective optimization,'' in \emph{Proceedings of the 22nd IEEE
  Congress on Evolutionary Computation (CEC'20)}, 2020, pp. 1--8.

\bibitem{Yu:2022:ECS}
X.~Yu, L.~Zhu, Y.~Wang, D.~Filev, and X.~Yao, ``Internal combustion engine
  calibration using optimization algorithms,'' \emph{Applied Energy}, vol. 305,
  p. 117894, 2022.

\bibitem{Chen:2019:Closer}
W.-Y. Chen, Y.-C. Liu, Z.~Kira, Y.-C.~F. Wang, and J.-B. Huang, ``A closer look
  at few-shot classification,'' in \emph{Proceedings of the 7th International
  Conference on Learning Representations (ICLR'19)}, 2019.

\bibitem{Wang:2020:GFA}
Y.~Wang, Q.~Yao, J.~T. Kwok, and L.~M. Ni, ``Generalizing from a few examples:
  A survey on few-shot learning,'' \emph{ACM Computing Surveys}, vol.~53,
  no.~3, pp. 1--34, 2020.

\bibitem{Hospedales:2021:Survey}
T.~M. Hospedales, A.~Antoniou, P.~Micaelli, and A.~J. Storkey, ``Meta-learning
  in neural networks: A survey,'' \emph{IEEE Transactions on Pattern Analysis
  and Machine Intelligence}, 2021.

\bibitem{Gupta:2017:TOS}
A.~Gupta, Y.-S. Ong, and L.~Feng, ``Insights on transfer optimization: Because
  experience is the best teacher,'' \emph{IEEE Transactions on Emerging Topics
  in Computational Intelligence}, vol.~2, no.~1, pp. 51--64, 2017.

\bibitem{Ding:2017:GMF}
J.~Ding, C.~Yang, Y.~Jin, and T.~Chai, ``Generalized multitasking for
  evolutionary optimization of expensive problems,'' \emph{IEEE Transactions on
  Evolutionary Computation}, vol.~23, no.~1, pp. 44--58, 2017.

\bibitem{Liaw:2019:EMO}
R.-T. Liaw and C.-K. Ting, ``Evolutionary manytasking optimization based on
  symbiosis in biocoenosis,'' in \emph{Proceedings of the 33rd AAAI Conference
  on Artificial Intelligence (AAAI'19)}, 2019, pp. 4295--4303.

\bibitem{Zhang:2021:ASO}
L.~Zhang, Y.~Xie, J.~Chen, L.~Feng, C.~Chen, and K.~Liu, ``A study on multiform
  multi-objective evolutionary optimization,'' \emph{Memetic Computing},
  vol.~13, no.~3, pp. 307--318, 2021.

\bibitem{Guo:2022:GMBO}
Z.~Guo, H.~Liu, Y.-S. Ong, X.~Qu, Y.~Zhang, and J.~Zheng, ``Generative
  multiform bayesian optimization,'' \emph{IEEE Transactions on Cybernetics
  (Early access)}, 2022.

\bibitem{Volpp:2020:MetaBO}
M.~Volpp, L.~P. Fr{\"{o}}hlich, K.~Fischer, A.~Doerr, S.~Falkner, F.~Hutter,
  and C.~Daniel, ``Meta-learning acquisition functions for transfer learning in
  bayesian optimization,'' in \emph{Proceedings of the 8th International
  Conference on Learning Representations (ICLR'20)}, 2020.

\bibitem{Zhuang:2020:Review-TL}
F.~Zhuang, Z.~Qi, K.~Duan, D.~Xi, Y.~Zhu, H.~Zhu, H.~Xiong, and Q.~He, ``A
  comprehensive survey on transfer learning,'' \emph{Proceedings of the IEEE},
  vol. 109, no.~1, pp. 43--76, 2020.

\bibitem{Wistuba:2021:FSBO}
M.~Wistuba and J.~Grabocka, ``Few-shot bayesian optimization with deep kernel
  surrogates,'' in \emph{Proceedings of the 9th International Conference on
  Learning Representations (ICLR'21)}, 2021.

\bibitem{Patacchiola:2020:DKT}
M.~Patacchiola, J.~Turner, E.~J. Crowley, M.~O'Boyle, and A.~Storkey,
  ``Bayesian meta-learning for the few-shot setting via deep kernels,'' in
  \emph{Advance in Neural Information Processing Systems 33 (NeurIPS'20)},
  2020.

\bibitem{Wilson:2016:DKL}
A.~G. Wilson, Z.~Hu, R.~Salakhutdinov, and E.~P. Xing, ``Deep kernel
  learning,'' in \emph{Proceedings of the 19th International Conference on
  Artificial Intelligence and Statistics (AISTATS'16)}, 2016, pp. 370--378.

\bibitem{Rasmussen:2006:GP}
C.~K. Williams and C.~E. Rasmussen, \emph{Gaussian Processes for Machine
  Learning}.\hskip 1em plus 0.5em minus 0.4em\relax Cambridge, MA: MIT press,
  2006.

\bibitem{Stein:1999:Kriging}
M.~L. Stein, \emph{Interpolation of Spatial Data: Some Theory for
  Kriging}.\hskip 1em plus 0.5em minus 0.4em\relax New York, NY: Springer
  Science \& Business Media, 1999.

\bibitem{Sacks:1989:DACE}
J.~Sacks, W.~J. Welch, T.~J. Mitchell, and H.~P. Wynn, ``Design and analysis of
  computer experiments,'' \emph{Statistical Science}, vol.~4, no.~4, pp.
  409--423, 1989.

\bibitem{Tossou:2019:ADKL}
P.~Tossou, B.~Dura, F.~Laviolette, M.~Marchand, and A.~Lacoste, ``Adaptive deep
  kernel learning,'' \emph{arXiv preprint arXiv:1905.12131}, 2019.

\bibitem{Garnelo:2018:Neural}
M.~Garnelo, J.~Schwarz, D.~Rosenbaum, F.~Viola, D.~J. Rezende, S.~Eslami, and
  Y.~W. Teh, ``Neural processes,'' \emph{arXiv preprint arXiv:1807.01622},
  2018.

\bibitem{Jones:1998:EGO}
D.~R. Jones, M.~Schonlau, and W.~J. Welch, ``Efficient global optimization of
  expensive black-box functions,'' \emph{Journal of Global Optimization},
  vol.~13, no.~4, pp. 455--492, 1998.

\bibitem{Finn:2017:MAML}
C.~Finn, P.~Abbeel, and S.~Levine, ``Model-agnostic meta-learning for fast
  adaptation of deep networks,'' in \emph{Proceedings of the 34th International
  Conference on Machine Learning (ICML'17)}, 2017, pp. 1126--1135.

\bibitem{Harrison:2018:ALPaCA}
J.~Harrison, A.~Sharma, and M.~Pavone, ``Meta-learning priors for efficient
  online bayesian regression,'' in \emph{Proceedings of the 13th Workshop on
  the Algorithmic Foundations of Robotics (WAFR'18)}, 2018, pp. 318--337.

\bibitem{Zhu:2020:ECU}
L.~Zhu, Y.~Wang, A.~Pal, and G.~Zhu, ``Engine calibration using global
  optimization methods with customization,'' SAE Technical Paper, Tech. Rep.
  2020-01-0270, 2020.

\bibitem{Deb:2005:DTLZ}
K.~Deb, L.~Thiele, M.~Laumanns, and E.~Zitzler, ``Scalable test problems for
  evolutionary multiobjective optimization,'' in \emph{Evolutionary
  Multiobjective Optimization}.\hskip 1em plus 0.5em minus 0.4em\relax London,
  U.K.: Springer, 2005, pp. 105--145.

\bibitem{Pan:2018:CSEA}
L.~Pan, C.~He, Y.~Tian, H.~Wang, X.~Zhang, and Y.~Jin, ``A classification-based
  surrogate-assisted evolutionary algorithm for expensive many-objective
  optimization,'' \emph{IEEE Transactions on Evolutionary Computation},
  vol.~23, no.~1, pp. 74--88, 2018.

\bibitem{Song:2021:KTA2}
Z.~Song, H.~Wang, C.~He, and Y.~Jin, ``A kriging-assisted two-archive
  evolutionary algorithm for expensive many-objective optimization,''
  \emph{IEEE Transactions on Evolutionary Computation}, vol.~25, no.~6, pp.
  1013--1027, 2021.

\bibitem{Zhang:2010:MOEADEGO}
Q.~Zhang, W.~Liu, E.~Tsang, and B.~Virginas, ``Expensive multiobjective
  optimization by {MOEA/D} with gaussian process model,'' \emph{IEEE
  Transactions on Evolutionary Computation}, vol.~14, no.~3, pp. 456--474,
  2010.

\bibitem{Knowles:2006:ParEGO}
J.~Knowles, ``{ParEGO}: A hybrid algorithm with on-line landscape approximation
  for expensive multiobjective optimization problems,'' \emph{IEEE Transactions
  on Evolutionary Computation}, vol.~10, no.~1, pp. 50--66, 2006.

\bibitem{Chugh:2016:K-RVEA}
T.~Chugh, Y.~Jin, K.~Miettinen, J.~Hakanen, and K.~Sindhya, ``A
  surrogate-assisted reference vector guided evolutionary algorithm for
  computationally expensive many-objective optimization,'' \emph{IEEE
  Transactions on Evolutionary Computation}, vol.~22, no.~1, pp. 129--142,
  2016.

\bibitem{Yu:2019:OREA}
X.~Yu, X.~Yao, Y.~Wang, L.~Zhu, and D.~Filev, ``Domination-based ordinal
  regression for expensive multi-objective optimization,'' in \emph{Proceedings
  of the 2019 IEEE Symposium Series on Computational Intelligence (SSCI'19)},
  2019, pp. 2058--2065.

\bibitem{Tian:2017:PlatEMO}
Y.~Tian, R.~Cheng, X.~Zhang, and Y.~Jin, ``{PlatEMO}: A {MATLAB} platform for
  evolutionary multi-objective optimization [educational forum],'' \emph{IEEE
  Computational Intelligence Magazine}, vol.~12, no.~4, pp. 73--87, 2017.

\bibitem{McKay:2000:LHS}
M.~D. McKay, R.~J. Beckman, and W.~J. Conover, ``A comparison of three methods
  for selecting values of input variables in the analysis of output from a
  computer code,'' \emph{Technometrics}, vol.~42, no.~1, pp. 55--61, 2000.

\bibitem{Ishibuchi:2015:IGD+}
H.~Ishibuchi, H.~Masuda, Y.~Tanigaki, and Y.~Nojima, ``Modified distance
  calculation in generational distance and inverted generational distance,'' in
  \emph{Proceedings of the 8th International Conference on Evolutionary
  Multi-criterion Optimization (EMO'15)}, 2015, pp. 110--125.

\bibitem{Ma:2013:COE}
H.~Ma, ``Control oriented engine modeling and engine multi-objective optimal
  feedback control,'' Ph.D. dissertation, University of Birmingham, 2013.

\end{thebibliography}


\begin{thebibliography}{1}
\providecommand{\url}[1]{#1}
\csname url@samestyle\endcsname
\providecommand{\newblock}{\relax}
\providecommand{\bibinfo}[2]{#2}
\providecommand{\BIBentrySTDinterwordspacing}{\spaceskip=0pt\relax}
\providecommand{\BIBentryALTinterwordstretchfactor}{4}
\providecommand{\BIBentryALTinterwordspacing}{\spaceskip=\fontdimen2\font plus
\BIBentryALTinterwordstretchfactor\fontdimen3\font minus
  \fontdimen4\font\relax}
\providecommand{\BIBforeignlanguage}[2]{{%
\expandafter\ifx\csname l@#1\endcsname\relax
\typeout{** WARNING: IEEEtran.bst: No hyphenation pattern has been}%
\typeout{** loaded for the language `#1'. Using the pattern for}%
\typeout{** the default language instead.}%
\else
\language=\csname l@#1\endcsname
\fi
#2}}
\providecommand{\BIBdecl}{\relax}
\BIBdecl

\bibitem{Stein:1999:Kriging}
M.~L. Stein, \emph{Interpolation of Spatial Data: Some Theory for
  Kriging}.\hskip 1em plus 0.5em minus 0.4em\relax New York, NY: Springer
  Science \& Business Media, 1999.

\bibitem{Finn:2017:MAML}
C.~Finn, P.~Abbeel, and S.~Levine, ``Model-agnostic meta-learning for fast
  adaptation of deep networks,'' in \emph{Proceedings of the 34th International
  Conference on Machine Learning (ICML'17)}, 2017, pp. 1126--1135.

\bibitem{Harrison:2018:ALPaCA}
J.~Harrison, A.~Sharma, and M.~Pavone, ``Meta-learning priors for efficient
  online bayesian regression,'' in \emph{Proceedings of the 13th Workshop on
  the Algorithmic Foundations of Robotics (WAFR'18)}, 2018, pp. 318--337.

\bibitem{Patacchiola:2020:DKT}
M.~Patacchiola, J.~Turner, E.~J. Crowley, M.~O'Boyle, and A.~Storkey,
  ``Bayesian meta-learning for the few-shot setting via deep kernels,'' in
  \emph{Advance in Neural Information Processing Systems 33 (NeurIPS'20)},
  2020.

\bibitem{Knowles:2006:ParEGO}
J.~Knowles, ``{ParEGO}: A hybrid algorithm with on-line landscape approximation
  for expensive multiobjective optimization problems,'' \emph{IEEE Transactions
  on Evolutionary Computation}, vol.~10, no.~1, pp. 50--66, 2006.

\bibitem{Yu:2019:OREA}
X.~Yu, X.~Yao, Y.~Wang, L.~Zhu, and D.~Filev, ``Domination-based ordinal
  regression for expensive multi-objective optimization,'' in \emph{Proceedings
  of the 2019 IEEE Symposium Series on Computational Intelligence (SSCI'19)},
  2019, pp. 2058--2065.

\end{thebibliography}

\end{document}


%
\title{Supplementary Material to ``Experience-Based Evolutionary Algorithms for Expensive Optimization"}

\author{Xunzhao~Yu, 
        Yan~Wang, 
        Ling~Zhu,
        Dimitar~Filev,~\IEEEmembership{Fellow,~IEEE}
        and~Xin~Yao,~\IEEEmembership{Fellow,~IEEE}
\thanks{This work was supported by Ford USA through a research grant to the University of Birmingham, UK. 
Xin Yao was also supported by the European Union's Horizon 2020 research and innovation programme under grant agreement number 766186 (ECOLE),  the Guangdong Provincial Key Laboratory (Grant No. 2020B121201001), the Program for Guangdong Introducing Innovative and Enterpreneurial Teams (Grant No. 2017ZT07X386), and Shenzhen Science and Technology Program (Grant No. KQTD2016112514355531).}
\thanks{
	X. Yu and X. Yao (the corresponding author) are with the CERCIA, 
	School of Computer Science, 
	University of Birmingham, Birmingham, B15 2TT, U.K. 
	(e-mail: xxy653@cs.bham.ac.uk; x.yao@cs.bham.ac.uk).
	X. Yao is also with the Research Institute of Trustworthy Autonomous Systems, and Guangdong Provincial Key Laboratory of Brain-inspired Computation, Department of Computer Science and Engineering, Southern University of Science and Technology, Shenzhen, China.}
\thanks{Y. Wang, L. Zhu, and D. Filev are with the Ford Motor Company, 
	2101 Village Rd, Dearborn, MI 48124, USA. 
	(e-mail: ywang21@ford.com, lzhu40@ford.com, dfilev@ford.com)}
}

\markboth{}
{Yu \MakeLowercase{\textit{et al.}}: Supplementary Material - Experience-based Evolutionary Algorithms for Expensive Optimization}

\maketitle

\IEEEpeerreviewmaketitle

\section{Generation of DTLZ variants}
Our DTLZ optimization experiments generate $M$-objective DTLZ variants in the following ways: \\
\textbf{DTLZ1}:
\begin{equation} \label{equ: DTLZ1-1}
	f_1 = (a_1 + g) 0.5 \prod_{i=1}^{M-1} y_i,
\end{equation}
\begin{equation} \label{equ: DTLZ1-2}
	f_{m = 2 : M - 1} = (a_m + g) (0.5 \prod_{i=1}^{M-m} y_i) (1 - y_{M-m+1}),
\end{equation}
\begin{equation} \label{equ: DTLZ1-3}
	f_M = (a_M + g) 0.5 (1 - y_1),
\end{equation}
\begin{equation}  \label{equ: DTLZ1-4}
	g = 100 \left[k + \sum_{i=1}^k \left( (z_i - 0.5)^2 - cos\left(20\pi (z_i - 0.5) \right) \right) \right],
\end{equation}
where $\textbf{y}$ is a vector that consists of the first $M$-1 variables in D-dimensional decision vector $\textbf{x}$. 
$\textbf{z}$ is a vector consisting of the rest $k$=$D$-$M$+1 variables in $\textbf{x}$. 
The variants of DTLZ1 introduce only one variable $\textbf{a} \in [0.1, 5.0]^M$ in Eq.(\ref{equ: DTLZ1-1}), Eq.(\ref{equ: DTLZ1-2}), and Eq.(\ref{equ: DTLZ1-3}), where $\textbf{a}=\textbf{1}$ in the original DTLZ1. For out-of-range test, $\textbf{a} \in [1.5, 5.0]^M$.
\quad \\

\textbf{DTLZ2}:
\begin{equation} \label{equ: DTLZ2-1}
	f_1 = (a_1 + g) \prod_{i=1}^{M-1} cos (\frac{y_i \pi}{b_1}),
\end{equation}
\begin{equation} \label{equ: DTLZ2-2}
	f_{m = 2 : M - 1} = (a_m + g) \left(\prod_{i=1}^{M-m} cos (\frac{y_i \pi}{b_m}) \right) sin (\frac{y_{M-m+1} \pi}{b_m}),
\end{equation}
\begin{equation} \label{equ: DTLZ2-3}
	f_M = (a_M + g) sin (\frac{y_1 \pi}{b_M}),
\end{equation}
\begin{equation} \label{equ: DTLZ2-4}
	g = \sum_{i=1}^k (z_i - 0.5)^2.
\end{equation} 
The variants of DTLZ2 introduces two variables $\textbf{a} \in [0.1, 5.0]^M$ and $\textbf{b} \in [0.5, 2.0]^M$ in Eq.(\ref{equ: DTLZ2-1}), Eq.(\ref{equ: DTLZ2-2}), and Eq.(\ref{equ: DTLZ2-3}), where $\textbf{a}=\textbf{1}$ and $\textbf{b}=\textbf{2}$ in the original DTLZ2. For out-of-range test, $\textbf{a} \in [1.5, 5.0]^M, \textbf{b} \in [0.5, 1.5]^M$.
\quad \\

\textbf{DTLZ3}:
The variants of DTLZ3 are generated using the same way as described in DTLZ2, except the equation $g$ from Eq.(\ref{equ: DTLZ2-4}) is replaced by the one from Eq.(\ref{equ: DTLZ1-4}).
\quad \\

\textbf{DTLZ4}:
The variants of DTLZ4 are generated using the same way as described in DTLZ2, except all $y_i$ are replaced by $y_i^{100}$.
\quad \\

\textbf{DTLZ5}:
The variants of DTLZ5 are generated using the same way as described in DTLZ2, except all $y_2, \dots, y_{M-1}$ are replaced by $\frac{1 + 2gy_i}{2 (1 + g)}$.
\quad \\

\textbf{DTLZ6}:
\begin{equation} \label{equ: DTLZ6}
	g = \sum_{i=1}^k z_i^{0.1}.
\end{equation}
The variants of DTLZ6 are generated using the same way as described in DTLZ5, except the equation $g$ from Eq.(\ref{equ: DTLZ2-4}) is replaced by the one from Eq.(\ref{equ: DTLZ6}).
\quad \\

\textbf{DTLZ7}:
\begin{equation} \label{equ: DTLZ7-1}
	f_{m = 1 : M-1} = y_m + a_m,
\end{equation}
\begin{equation} \label{equ: DTLZ7-2}
	f_M = (1 + g) \left( M - \sum_{i=1}^{M-1} \left[\frac{f_i}{1 + g} \left(1 + sin(3 \pi f_i) \right) \right] \right),
\end{equation}
\begin{equation} \label{equ: DTLZ7-3}
	g = a_M + 9 \sum_{i=1}^k \frac{z_i}{k}.
\end{equation}
The variants of DTLZ7 introduce two variables $\textbf{a} \in [0.1, 5.0]^M$ in Eq.(\ref{equ: DTLZ7-1}) and Eq.(\ref{equ: DTLZ7-3}), where $a_{m=1:M-1}=0$ and $a_M=1$ in the original DTLZ7. For out-of-range test, $\textbf{a} \in [1.5, 5.0]^M$.

\section{Effectiveness of Learning Experiences: Sinusoid Function Regression}
\begin{table*}[!t]
	\caption[Average NMSE and standard deviation (in brackets) of 30 runs on the amplitude regression of sinusoid function.]
	{Average NMSE and standard deviation (in brackets) of 30 runs on the amplitude regression of sinusoid function. 
	GP \cite{Stein:1999:Kriging} is a widely used surrogate in SAEAs, MAML \cite{Finn:2017:MAML}, ALPaCA \cite{Harrison:2018:ALPaCA}, and DKT \cite{Patacchiola:2020:DKT}  are meta-learning methods. GP\_Adam is a GP model fitted by Adam optimizer. DKL is a deep kernel learning algorithm that adds a neural network to GP\_Adam. MDKL\_NN applies meta-learning to DKL, but no common base kernel parameters are shared between related tasks. 
	Support data points are used to train non-meta surrogates or adapt meta-learning surrogates.
	The symbols '$+$', '$\approx$', '$-$' denote the win/tie/loss result of Wilcoxon rank sum test (significance level is 0.05) between MDKL and comparison modeling methods, respectively. The last row counts the total win/tie/loss results.}
	\label{tab: sinusoid}
	\centering
	\scalebox{0.80}{ 
	\begin{tabular}{l| c| c c c c| c c c}
	\hline
	Support data	& GP \cite{Stein:1999:Kriging} 	& GP\_Adam 			&	DKL 					& 	MDKL\_NN 			&	MDKL		&	DKT \cite{Patacchiola:2020:DKT}	&	MAML \cite{Finn:2017:MAML}	& ALPaCA \cite{Harrison:2018:ALPaCA} \\
	points $|S_*|$ 	&						&					&						&						&				&						&						& \\
	\hline 
        	2			& 1.63e-1(9.18e-2)$\approx$ 	& 1.93e-1(9.72e-2)$+$	& 1.63e-1(9.05e-2)$\approx$	& 1.57e-1(9.26e-2)$\approx$ 	& 1.56e-1(9.49e-2) 	& 1.56e-1(9.49e-2)$\approx$ 	& 2.09e-1(3.63e-1)$\approx$	& 1.07e+0(2.57e+0)$\approx$ \\
        	3			& 1.27e-1(6.04e-2)$\approx$ 	& 1.62e-1(6.53e-2)$+$ 	& 1.21e-1(5.96e-2)$\approx$ 	& 1.16e-1(5.95e-2)$\approx$ 	& 1.10e-1(6.20e-2) 	& 1.10e-1(6.20e-2)$\approx$ 	& 2.09e-1(3.60e-1)$\approx$	& 4.36e-1(8.57e-1)$\approx$\\
        	5			& 6.76e-2(4.62e-2)$\approx$ 	& 1.09e-1(5.61e-2)$+$ 	& 7.52e-2(4.40e-2)$+$ 		& 6.38e-2(3.91e-2)$\approx$ 	& 4.79e-2(3.73e-2) 	& 4.79e-2(3.70e-2)$\approx$ 	& 2.08e-1(3.59e-1)$+$ 		& 4.31e-1(8.04e-1)$\approx$ \\
        	10			& 1.70e-2(1.87e-2)$\approx$ 	& 6.13e-2(4.58e-2)$+$ 	& 2.87e-2(1.89e-2)$+$ 		& 1.89e-2(1.61e-2)$+$ 		& 1.07e-2(1.16e-2) 	& 1.09e-2(1.17e-2)$\approx$ 	& 2.08e-1(3.58e-1)$+$ 		& 6.59e-1(2.14e+0)$+$ \\
        	20			& 5.42e-3(7.64e-3)$+$ 		& 3.92e-2(4.29e-2)$+$ 	& 9.64e-3(1.02e-2)$+$ 		& 5.24e-3(6.57e-3)$+$ 		& 2.57e-3(4.53e-3) 	& 2.63e-3(4.61e-3)$\approx$ 	& 2.08e-1(3.58e-1)$+$ 		& 1.13e-1(3.39e-1)$+$ \\
        	30			& 3.97e-3(7.40e-3)$+$ 		& 3.32e-2(4.18e-2)$+$	& 4.81e-3(6.68e-3)$+$		& 3.20e-3(5.85e-3)$+$ 		& 1.68e-3(3.61e-3) 	& 1.60e-3(3.39e-3)$\approx$ 	& 2.08e-1(3.58e-1)$+$ 		& 7.59e-2(2.01e-1)$+$ \\
	\hline 
	win/tie/loss 	& 2/4/0					& 6/0/0				& 4/2/0					& 3/3/0					& -/-/-			& 0/6/0					& 4/2/0  					& 3/3/0 \\
	\hline
  	\end{tabular}}
\end{table*}
In the sinusoid regression experiment, we learn experiences from a series of 1-dimensional sinusoid functions:
\begin{equation} \label{equ: sinusoid}
	y = A sin(wx + b) + \epsilon
\end{equation} 
where the amplitude $A$ and phase $w$ of sine waves are varied between functions. The target is to approximate an unknown sinusoid function with a small support dataset $S_*$ and the learned experiences. 
Clearly, by integrating experiences with $S_*$, we estimate parameters $(A, w, b)$ for an unknown sinusoid function. As a result, the output $y$ of the given sinusoid function can be predicted once a query data $x$ is inputted.

\subsection{Generation of Sinusoid Function Variants}
As suggested in \cite{Finn:2017:MAML, Harrison:2018:ALPaCA}, we set amplitude $A \in$ [0.1, 5.0], frequency $w \in [0.999, 1.0]$, phase $b \in$ [0, $\pi$], and Gaussian noise $\epsilon \sim(0, 0.1)$. 
Therefore, a sinusoid function can be generated by sampling three parameters $(A, w, b)$ from their ranges uniformly. In total, $N_m = N =$20000 related sinusoid functions are generated at random.

\subsection{Experimental Setups}
All data points $x$ are sampled from the range $\in$ [-5.0, 5.0]. In the meta-learning procedure, both support set and query set contain 5 data points. Hence, a dataset $D_i$ is sampled from each (related) sinusoid function $T_i$, and $|D_i|= |D_m|=10$. Six experiments are conducted where $|S_*|=\{2, 3, 5, 10, 20, 30\}$ data points are sampled from the target function. Considering Gaussian noise $\epsilon$ could be relatively large when amplitude $A$ is close to 0.1, normalized mean squared error (NMSE) is chosen as a performance indicator. NMSE is measured using a dataset that contains 100 data points sampled uniformly from the $x$ range.
The comparison algorithms are the same as described in Section IV-A of the main file. 

\subsection{Results and Analysis:}
Table S\ref{tab: sinusoid} reports the statistical test results of the NMSE values achieved by comparison algorithms on sinusoid function regression experiments. Each row lists the results obtained when the same number of fitness evaluations $|S_*|$ are used to train models.
The results of Wilcoxon rank sum test between MDKL and other compared algorithms are listed in the last row. 
It can be observed that both MDKL and DKT have achieved the smallest NMSE values on all tests in the comparison with other meta-learning and non-meta-learning modeling methods.

Contributions of MDKL components are analyzed through statistical tests between MDKL related methods.
The statistical test results between DKL and GP\_Adam are 5/1/0, showing that DKL is preferable to GP\_Adam when only a few data points are available for modeling. Hence, using a neural network to build a deep kernel for GP is able to enhance the performance of modeling.
When meta-learning technique is applied to DKL, the statistical test results between MDKL\_NN and DKL are 3/3/0. The meta-learning of neural network parameters is necessary since it contributes to the performance of MDKL.
Further statistical test between MDKL and MDKL\_NN gives results of 3/3/0, indicating that the meta-learning of base kernel parameters is effective on this regression problem.

\section{Performance on Expensive Multi-Objective Optimization: Result Figure}
\begin{figure*}[!t]	
	\centerline{
		\includegraphics[width=3.2in]{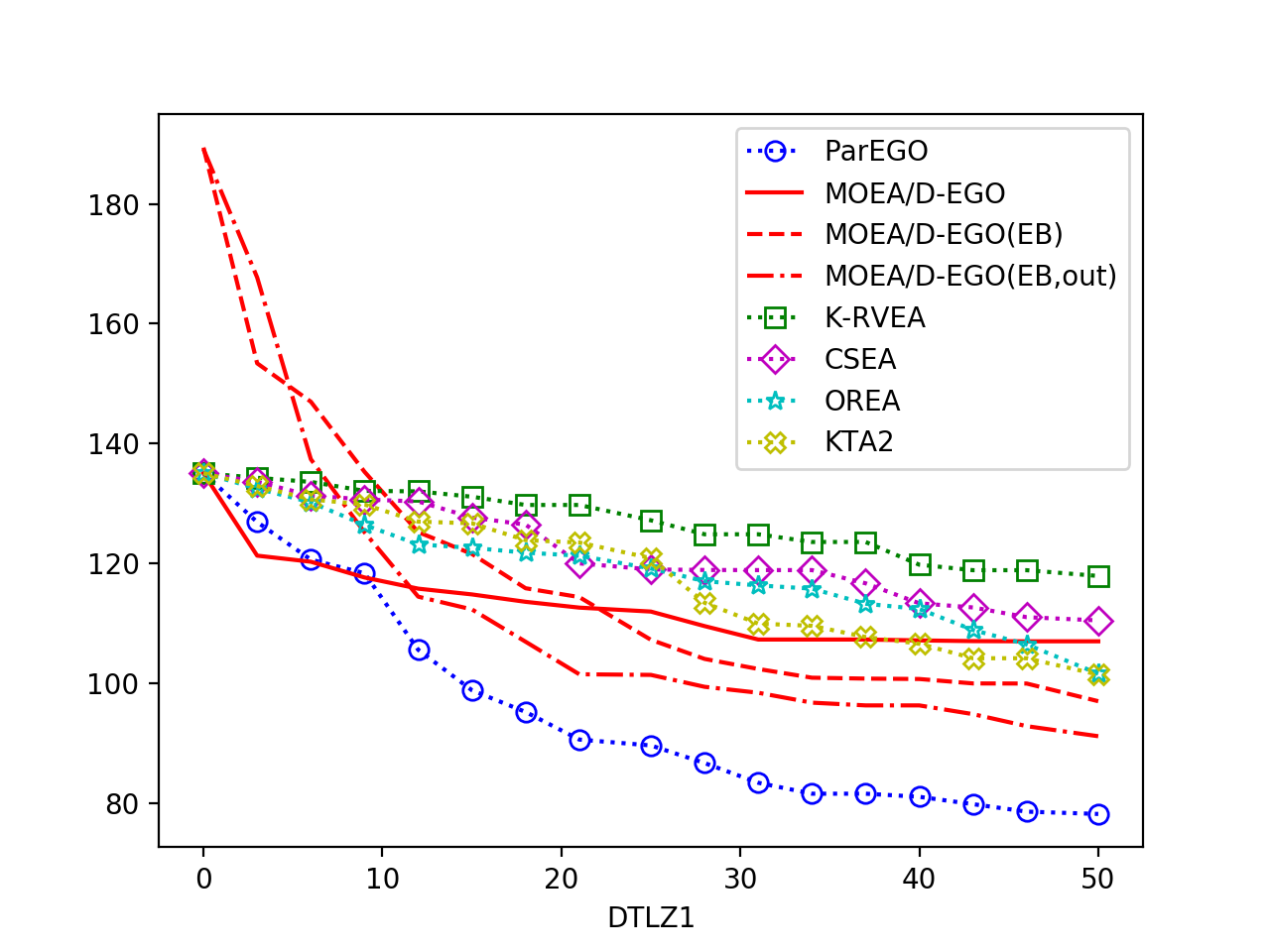}
		\includegraphics[width=3.2in]{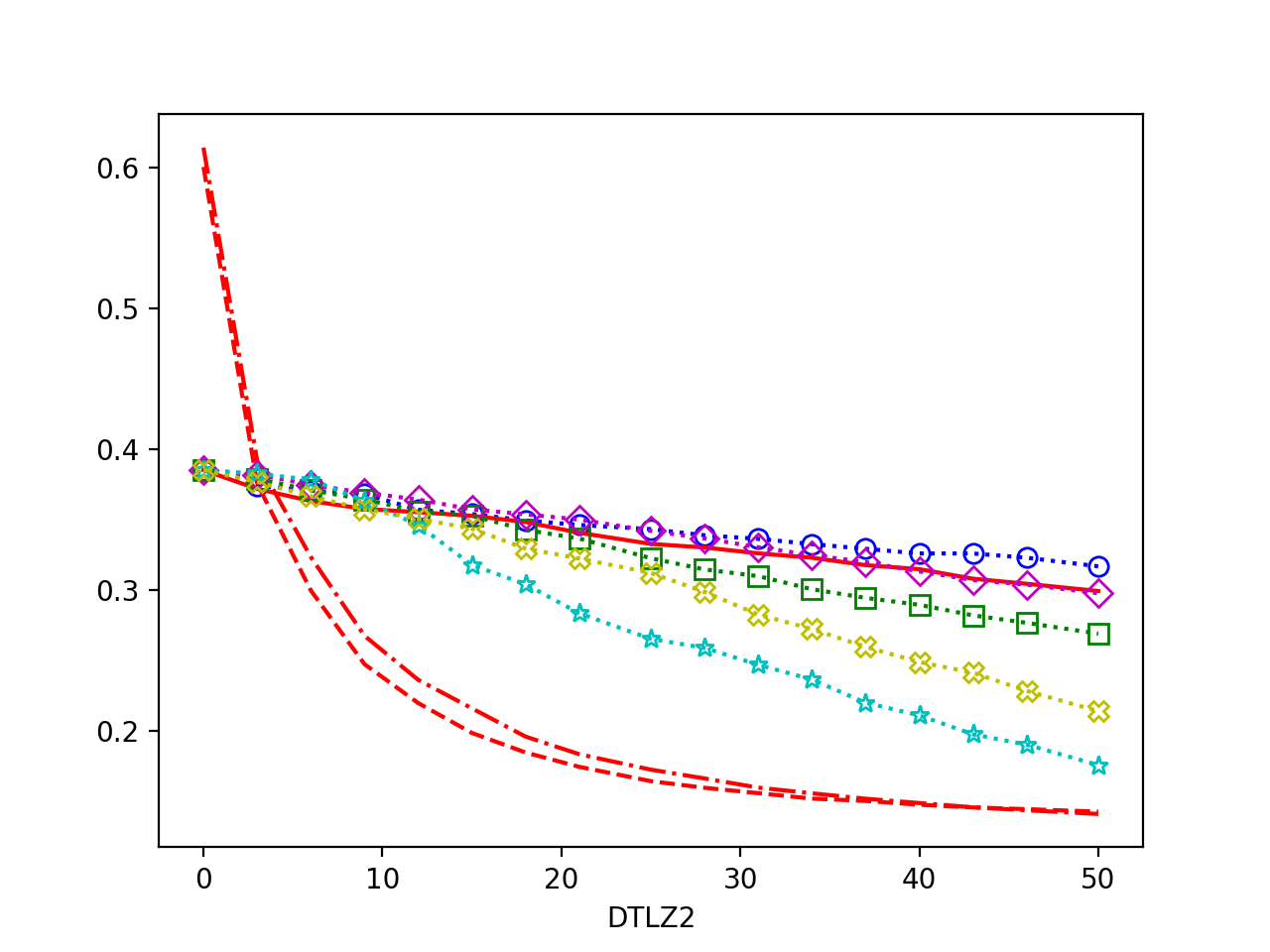}
	}
	\centerline{
		\includegraphics[width=3.2in]{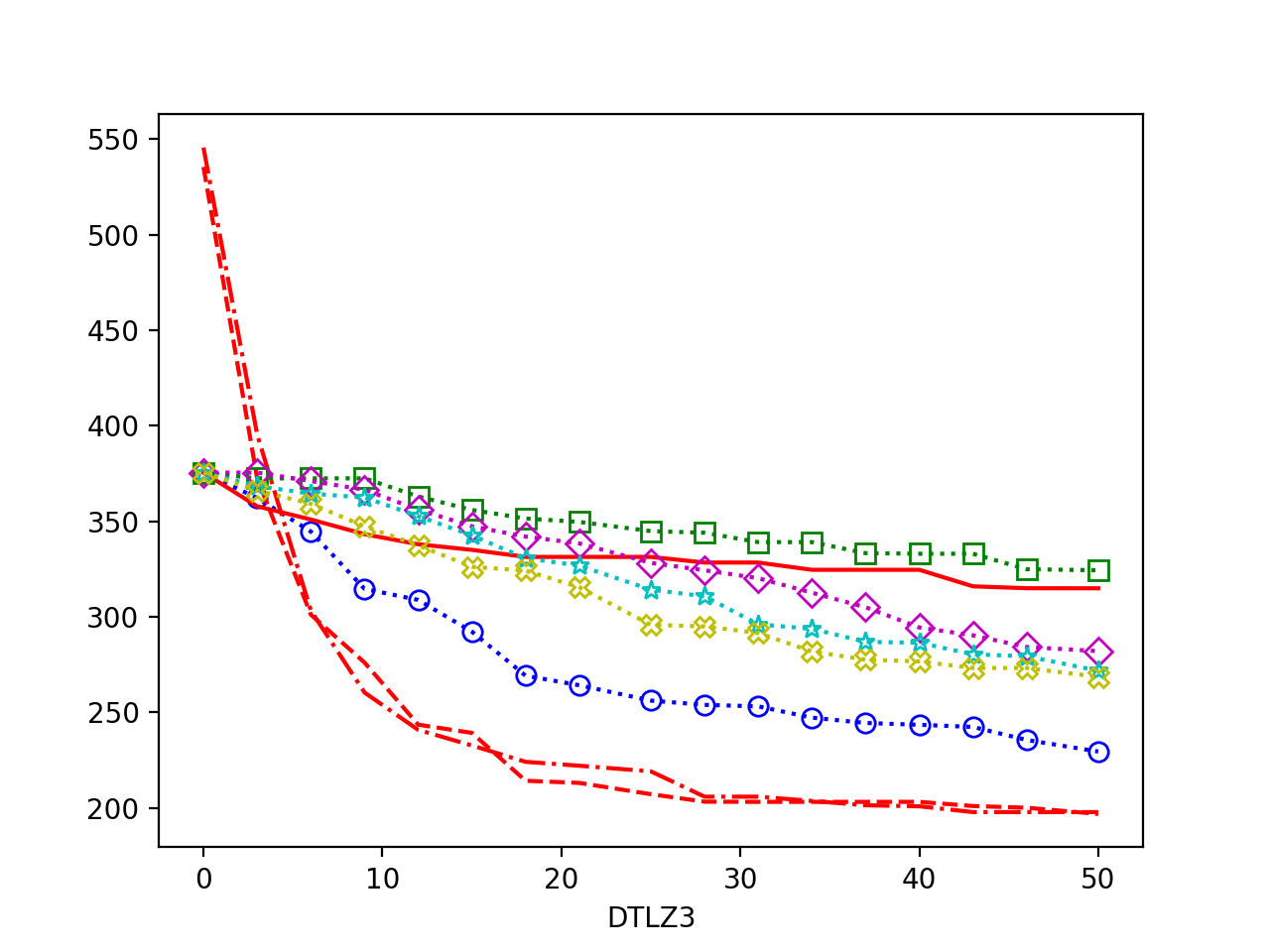} 
		\includegraphics[width=3.2in]{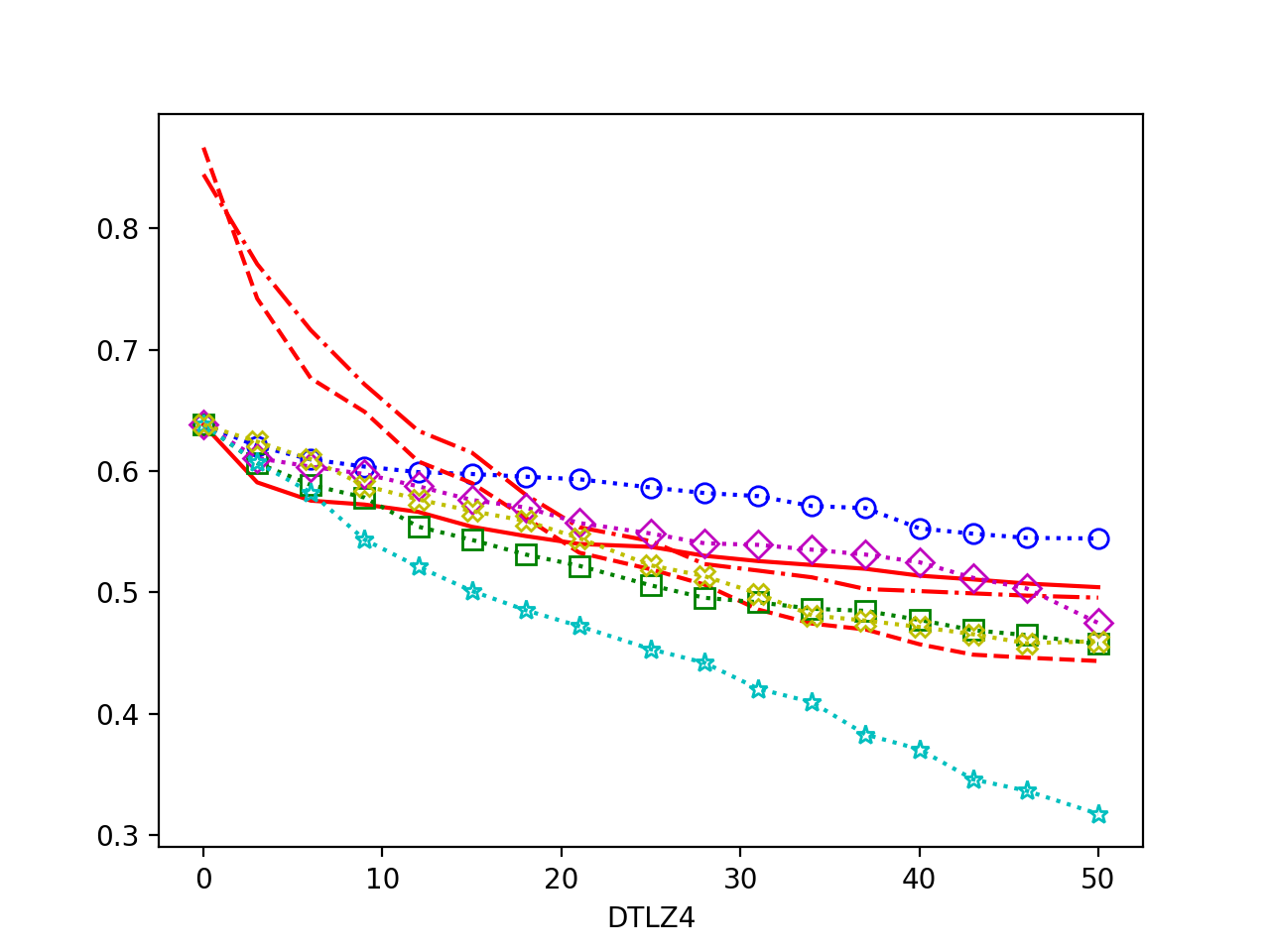} 
	}
	\centerline{
		\includegraphics[width=2.5in]{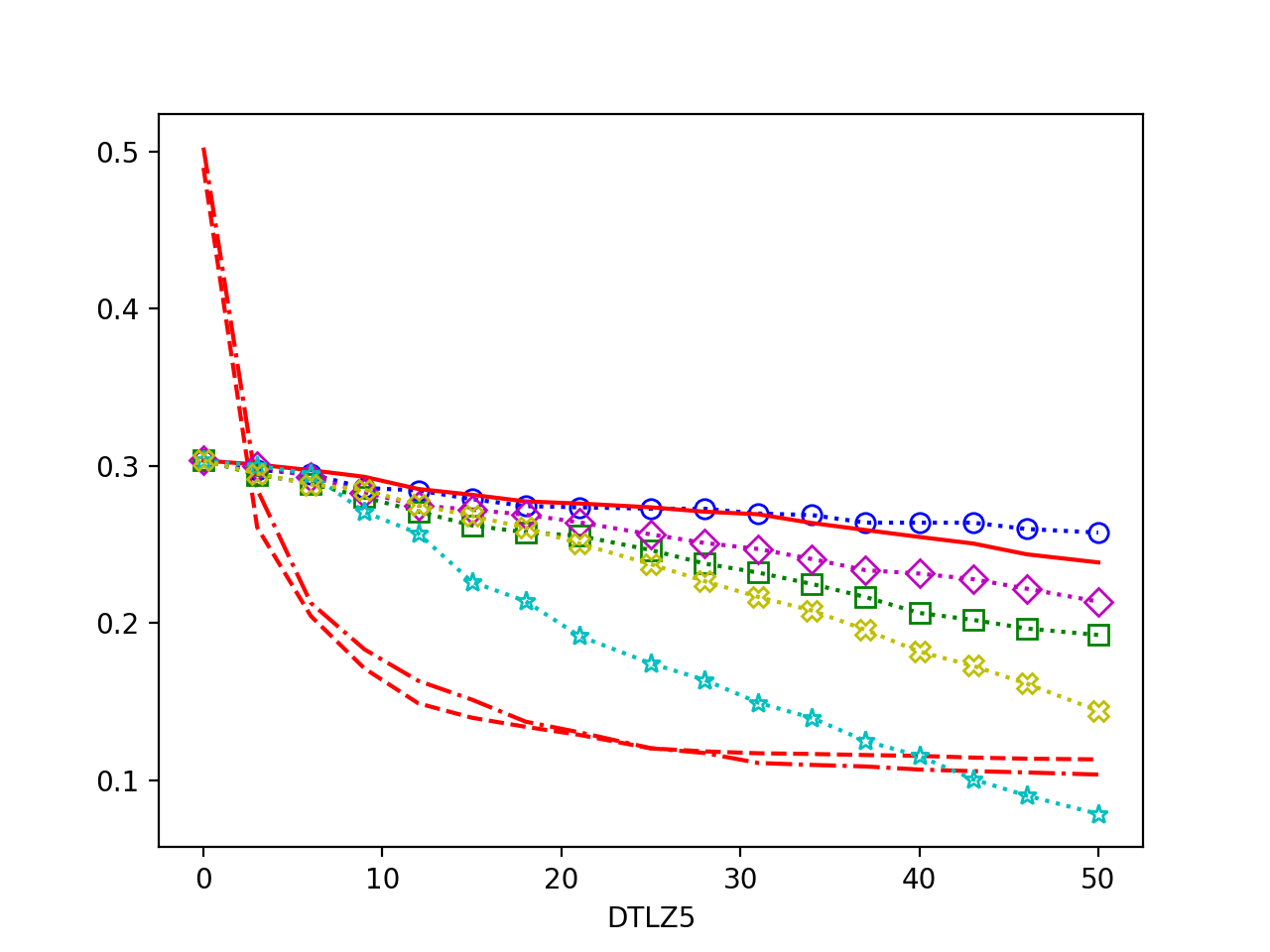} 
		\includegraphics[width=2.5in]{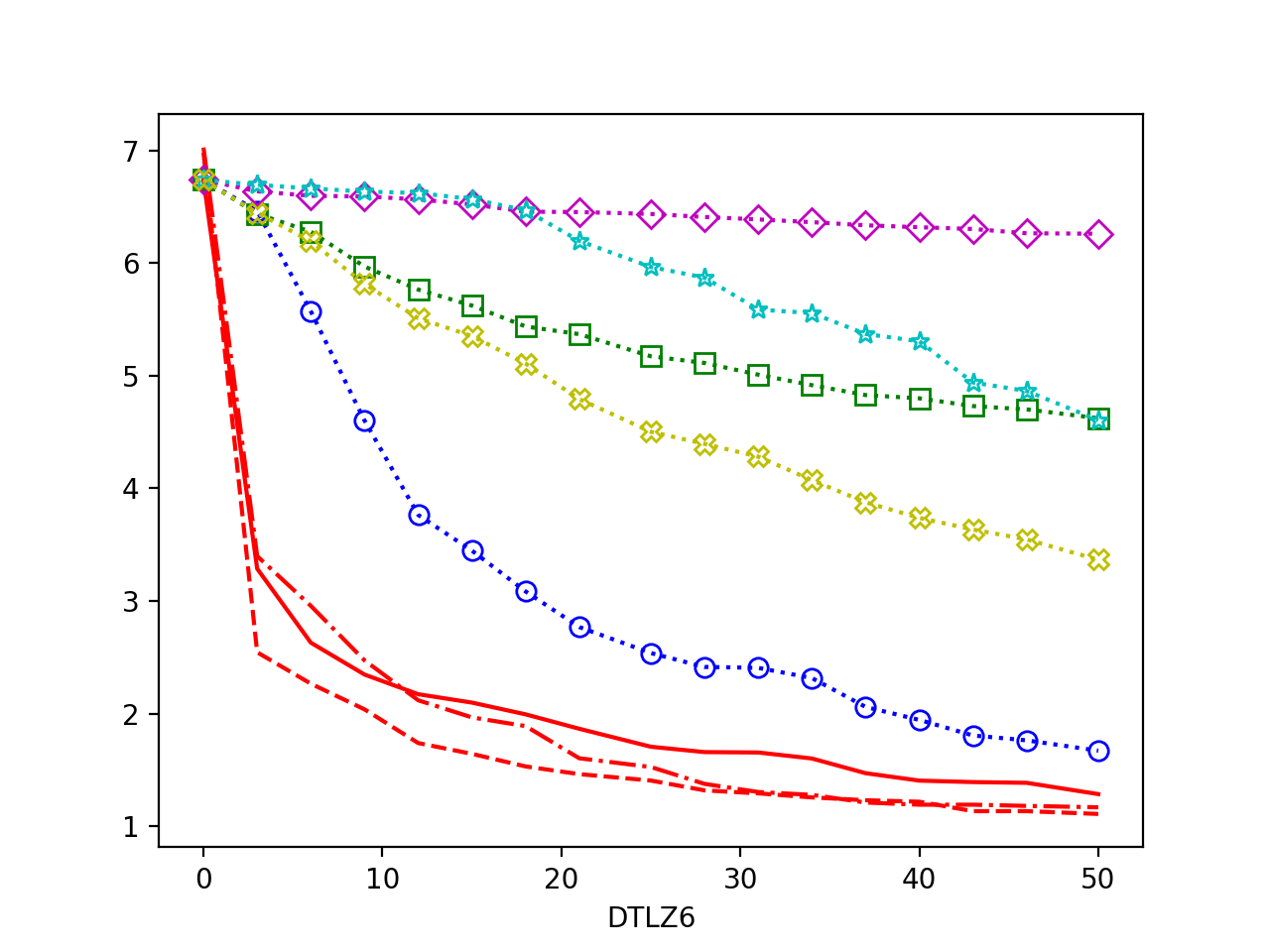} 
		\includegraphics[width=2.5in]{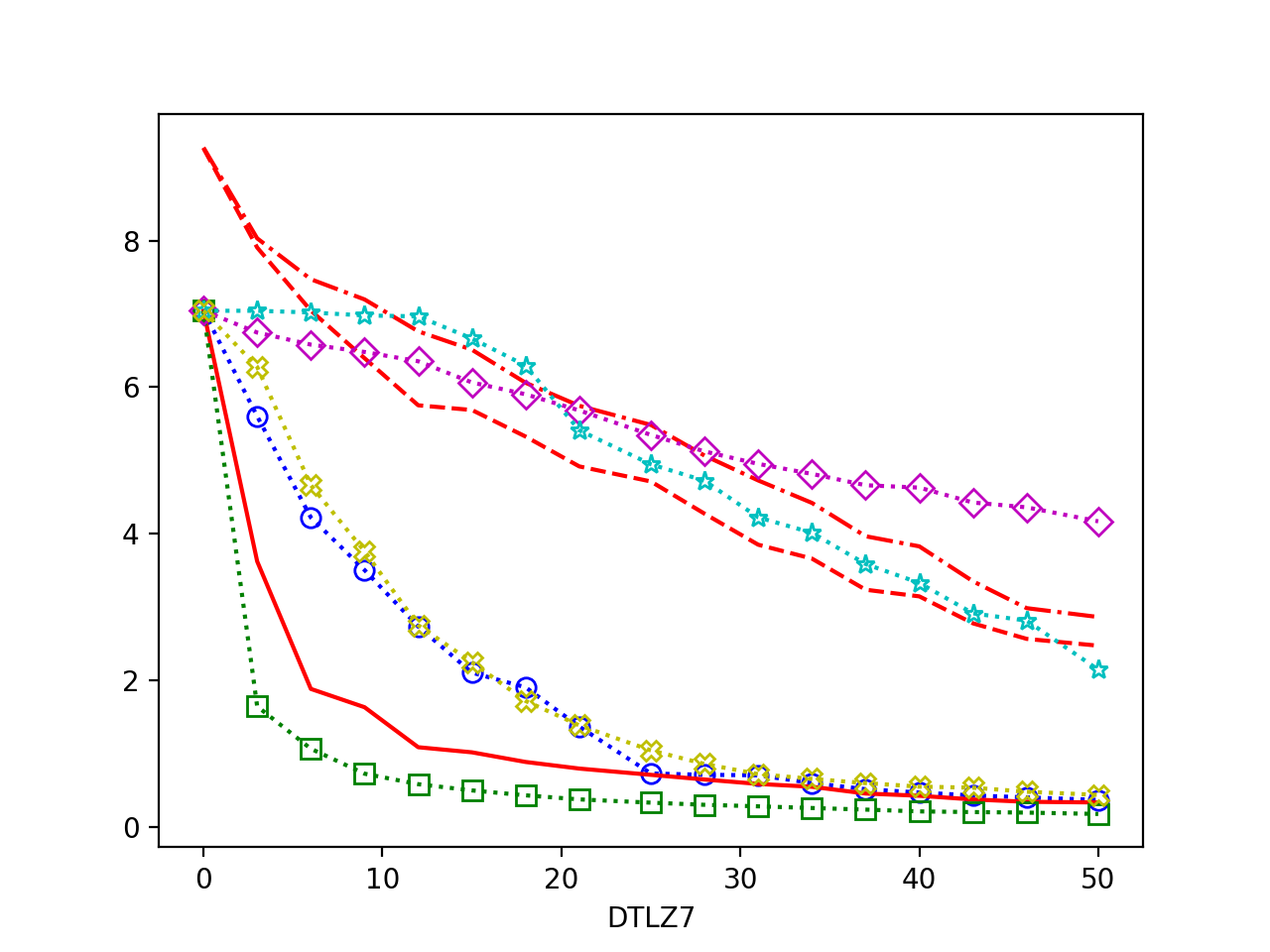} 
	}
	\caption[Average IGD+ values obtained from 30 runs on DTLZ functions. 50 evaluations for further optimization.]
	{Average IGD+ values obtained from 30 runs on DTLZ functions.
	MOEA/D-EGO(EB)s and comparison algorithms initialize their surrogates with 10, 100 samples, respectively. Extra 50 evaluations are allowed during the optimization process. 
	Note that 'EB out' indicates the target task is excluded from the range of related tasks during the MDKL meta-learning procedure.
	X-axis denotes the number of evaluations used after the surrogate initialization. 
	In comparison to MOEA/D-EGO, both MOEA/D-EGO(EB)s achieve smaller IGD+ values on DTLZ1, DTLZ2, DTLZ3, DTLZ5 and competitive IGD+ values on DTLZ4, DTLZ6, while 90 evaluations are saved by using experiences.}
	\label{fig: DTLZ150}
\end{figure*}

Fig. \ref{fig: DTLZ150} illustrates the results given by Tables 
IV and V.
MOEA/D-EGO(EB, out) refers to the results obtained in the `out-of-range' situation. As discussed in Sections 
IV-B and IV-C of the main file, when compared with MOEA/D-EGO, both MOEA/D-EGO(EB)s achieve smaller IGD+ values on DTLZ1, DTLZ2, DTLZ3, DTLZ5 and competitive IGD+ values on DTLZ4, DTLZ6, while 90 evaluations are saved by using experiences. 

\section{Performance on Expensive Multi-Objective Optimization Under the Same Evaluation Budget} \label{sec: exp-DTLZ10}
The statistical test results reported in the last row of Table 
IV show that ParEGO \cite{Knowles:2006:ParEGO} and OREA \cite{Yu:2019:OREA} are the best two comparison algorithms when compared with our MOEA/D-EGO(EB). In this section, we evaluate the performance of MOEA/D-EGO(EB) when no extra evaluation is saved.
	For this purpose, we compare the optimization performance of these three SAEAs under the same evaluation budget: 10 evaluations (1$d$) for surrogate initialization and 50 evaluations for further optimization. The statistical test results are reported in Table S\ref{tab: DTLZadditional}. 
\begin{table*}[!t]
	\caption[Average IGD+ values and standard deviation (in brackets) obtained from 30 runs on DTLZ functions. Same evaluation budget.]
	{Average IGD+ values and standard deviation (in brackets) obtained from 30 runs on DTLZ functions.
	 MOEA/D-EGO(EB) is compared with ParEGO and OREA under the same evaluation budget: 10 evaluations for surrogate initialization and 50 evaluations for the optimization process.
	The symbols '$+$', '$\approx$', '$-$' denote the win/tie/loss result of Wilcoxon rank sum test (significance level is 0.05) between MOEA/D-EGO(EB) and two comparison algorithms, respectively. The last row counts the total win/tie/loss results.
	}
	\label{tab: DTLZadditional}
	\centering
	\begin{tabular}{l c c c}
        \hline
    	Problem 		& MOEA/D-EGO(EB) 	& ParEGO 				& OREA 		\\
	\hline 
    	DTLZ1		& 9.70e+1(1.87e+1)		& 6.70e+1(4.75e+0)$-$ 		& 1.10e+2(3.65e+1)$\approx$ \\ 
	DTLZ2		& 1.43e-1(2.29e-2)		& 5.51e-1(5.37e-2)$+$		& 4.28e-1(6.68e-2)$+$	\\
	DTLZ3		& 1.97e+2	(1.64e+1)		& 1.84e+2(8.86e+0)$\approx$ 	& 2.72e+2(6.59e+1)$+$ 	\\ 
	DTLZ4		& 4.44e-1(1.35e-1)		& 6.29e-1(7.99e-2)$+$		& 6.45e-1(1.24e-1)$+$	\\
	DTLZ5		& 1.13e-1(2.24e-2) 		& 4.32e-1(8.88e-2)$+$		& 3.02e-1(7.63e-2)$+$	\\ 
	DTLZ6		& 1.11e+0(5.71e-1)		& 1.03e+0(4.78e-1)$\approx$ 	& 5.71e+0(6.73e-1)$+$	\\ 
	DTLZ7		& 2.47e+0(1.89e+0) 		& 4.38e-1(1.39e-1)$-$		& 7.12e+0(1.77e+0)$+$	\\ 
	\hline 
	win/tie/loss	& -/-/-				& 3/2/2					& 6/1/0  \\ 
	\hline
	\end{tabular}
\end{table*}
	It can be seen that our MOEA/D-EGO(EB) outperforms the compared SAEAs when only 1$d$ evaluations are used to initialize their surrogates. The effectiveness of our experience-based SAEA framework has been demonstrated.
	Note that OREA is an evolutionary algorithm assisted by ordinal-regression-based surrogates. Currently, our experience-based SAEA framework is applicable to the SAEAs working with regression-based surrogates. The meta-learning of ordinal-regression models can be a topic of further research.

\section{Experiments on Extremely Expensive Multi-Objective Optimization}
In this section, we investigate the performance of our SAEA framework in the context of extremely expensive optimization, where the allowed fitness evaluations on target problems are fewer than that in the experiment carried out in Sections 
IV-B and IV-C of the main file. 

\begin{table*}[!t]
	\caption[Average IGD+ values and standard deviation (in brackets) obtained from 30 runs on DTLZ functions. 30 evaluations for further optimization.]
	{Average IGD+ values and standard deviation (in brackets) obtained from 30 runs on DTLZ functions.
	MOEA/D-EGO(EB) and comparison algorithms initialize their surrogates with 10, 60 samples, respectively. Extra 30 evaluations are allowed during the optimization process.
	The symbols '$+$', '$\approx$', '$-$' denote the win/tie/loss result of Wilcoxon rank sum test (significance level is 0.05) between MOEA/D-EGO(EB) and comparison algorithms, respectively. The last row counts the total win/tie/loss results.
	Performance improvement can be observed from the comparisons between MOEA/D-EGO(EB) and MOEA/D-EGO, while 50 evaluations are saved from surrogate initialization.}
	\label{tab: DTLZ90}
	\centering
\scalebox{0.88}{
        \begin{tabular}{l| c c| c c c c c}
        \hline
        Problem	&	MOEA/D-EGO				& MOEA/D-EGO(EB)	& ParEGO 				& K-RVEA					& KTA2					& CSEA					& OREA	\\
        \hline
        DTLZ1		& 1.07e+2(2.73e+1)$\approx$	& 1.03e+2(2.34e+1)		& 8.70e+1(2.53e+1)$-$		& 1.22e+2(3.26e+1)$+$ 		& 1.15e+2	(3.18e+1)$\approx$ 	& 1.08e+2(2.68e+1)$\approx$	& 1.11e+2(2.25e+1)$+$ \\ 
        	DTLZ2		& 3.49e-1(5.82e-2)$+$		& 1.57e-1(2.29e-2)		& 3.51e-1(5.01e-2)$+$		& 3.72e-1(4.40e-2)$+$ 		& 3.57e-1(4.68e-2)$+$ 		& 3.55e-1(5.23e-2)$+$ 		& 3.14e-1(3.76e-2)$+$ \\
	DTLZ3		& 3.07e+2	(5.32e+1)$+$		& 2.03e+2(2.42e+1)		& 2.16e+2(4.89e+1)$\approx$	& 3.53e+2(7.89e+1)$+$ 		& 3.23e+2(8.82e+1)$+$ 		& 3.35e+2	(6.95e+1)$+$ 		& 3.39e+2(7.72e+1)$+$ \\ 
    	DTLZ4		& 5.45e-1(1.09e-1)$\approx$	& 4.91e-1(1.24e-1)		& 6.36e-1(8.67e-2)$+$		& 5.53e-1(9.96e-2)$\approx$	& 5.47e-1(1.04e-1)$\approx$	& 5.84e-1(9.75e-2)$+$ 		& 5.14e-1(1.21e-1)$\approx$ \\ 
	DTLZ5		& 2.79e-1 (5.69e-2)$+$		& 1.18e-1(2.25e-2) 		& 2.78e-1(5.59e-2)$+$		& 2.82e-1(5.52e-2)$+$		& 2.60e-1(5.59e-2)$+$ 		& 2.77e-1(4.41e-2)$+$		& 1.99e-1(4.53e-2)$+$ \\ 
    	DTLZ6		& 2.04e+0(7.33e-1)$+$		& 1.29e+0(6.44e-1)		& 2.47e+0(7.39e-1)$+$		& 5.23e+0(6.27e-1)$+$		& 4.58e+0(6.47e-1)$+$		& 6.44e+0 (3.59e-1)$+$ 		& 5.79e+0(6.70e-1)$+$ \\ 
    	DTLZ7		& 1.90e+0(9.19e-1)$-$  		& 4.16e+0(2.54e+0) 		& 1.39e+0(1.49e+0)$-$ 		& 3.13e-1(6.17e-2)$-$   		& 2.05e+0	(2.20e+0)$-$		& 5.47e+0(1.34e+0)$+$		& 5.51e+0(1.32e+0)$+$ \\         	  
	\hline  			
	win/tie/loss 	& 4/2/1					& -/-/-				& 4/1/2					& 5/1/1					& 4/2/1					& 6/1/0					& 6/1/0 \\  
	\hline  		
	\end{tabular}}
\end{table*}
\subsection{Performance between Comparison Algorithms} \label{sec: exp-DTLZ90}
We conduct the experiment described in Section 
IV-B of the main file, but with a smaller evaluation budget than the budget listed in Table 
III.
The size of the initial dataset $S_*$ is set to 10, 60 for our MOEA/D-EGO(EB) and comparison algorithms, respectively. 30 extra evaluations for further optimization are allowed. The total evaluation budget is 40, 90 for our MOEA/D-EGO(EB) and comparison algorithms, respectively.

The aim of this subsection is to answer the question below:
\begin{itemize}
	\item Is our experience-based SAEA framework more suitable for the optimization problems in which evaluations are extremely expensive? In other words, will the advantage of our SAEA framework become more prominent if the optimization problems allow a smaller evaluation budget? 
\end{itemize}

The comparison results reported in Table S\ref{tab: DTLZ90} show that MOEA/D-EGO(EB) has achieved competitive or smaller IGD+ values than MOEA/D-EGO on all DTLZ functions except for DTLZ7. Meanwhile, 5$d$ evaluations have been saved. 

	Consistent with the results discussed in Section 
IV-B of the main file, MOEA/D-EGO(EB) fails to achieve a competitive result compared to MOEA/D-EGO on DTLZ7 since experiences are learned from small datasets collected from related tasks. Although we set a different evaluation budget for all SAEAs, the size of datasets for meta-learning $|D_m|$ has not been modified. 
	However, it can be observed from the statistical test results (see the last row of Tables 
IV and S\ref{tab: DTLZ90}) that our MOEA/D-EGO(EB) outperforms the comparison algorithms on 26, 29 test instances when the total evaluation budget of comparison algorithms is set to 150, 90, respectively. 
	This answers the question we raised before: The advantage of our experience-based SAEA framework is more prominent in the extremely expensive problems where a smaller evaluation budget is allowed. 
	The comparison between the results obtained from Tables 
IV and S\ref{tab: DTLZ90} has demonstrated that our SAEA framework is preferable when solving optimization problems within a very limited evaluation budget.

\subsection{Out-Of-Range Analysis on Extremely Expensive Optimization}  \label{sec: exp-out90}
In Section 
IV-C of the main file, we carried out an experiment to study the influence of task similarity on the performance of experience-based expensive multi-objective optimization. The optimization results obtained from the `in-range' and the `out-of-range' situations are compared. 
In this subsection, we conduct an experiment to investigate the difference between the `in-range' and the `out-of-range' situations for extremely expensive multi-objective optimization. The experimental setups are the same as the setups described in Section 
IV-C of the main file, except the allowed fitness evaluation budget is the same as described in Section \ref{sec: exp-DTLZ90}.

Table S\ref{tab: DTLZ40-out} gives the statistical test results, it can be seen that the `out-of-range' situation achieves competitive IGD+ values to the `in-range' situation on all 7 test instances. 
In comparison to MOEA/D-EGO, the experience gained in the `out-of-range' situation leads to competitive or smaller IGD+ values on 6 DTLZ functions.
Furthermore, similar results can be observed in the last row of Table S\ref{tab: DTLZ40-out}, the `out-of-range' situation achieves better/competitive/worse IGD+ values than all compared SAEAs on 28/9/5 test instances. In comparison, the `in-range' situation achieves better/competitive/worse IGD+ values than all compared SAEAs on 29/8/5 test instances. There is only a minor difference between the optimization results obtained in two situations. 
These observations are consistent with the conclusions we made in Section 
IV-C of the main file.
\begin{table}[!t]
	\caption[Average IGD+ values and standard deviation (in brackets) obtained from 30 runs on DTLZ functions. Comparison between in-range and out-of-range, 30 evaluations for further optimization.]
	{Average IGD+ values and standard deviation (in brackets) obtained from 30 runs on DTLZ functions.
	Both MOEA/D-EGO(EB)s initialize their surrogates with 10 samples, extra 30 evaluations are allowed during the optimization process.
	`Out-of-range' indicates the target task is excluded from the range of related tasks during the MDKL meta-learning procedure.
	The symbols '$+$', '$\approx$', '$-$' denote the win/tie/loss result of Wilcoxon rank sum test (significance level is 0.05) between the `out-of-range' and the `in-range' situations, respectively.
	The last two rows count the statistical test results between MOEA/D-EGO(EB)s and other compared algorithms.}
	\label{tab: DTLZ40-out}
	\centering
        \begin{tabular}{l c c}
        \hline
        MOEA/D-EGO(EB)s		&	In-range				& Out-of-range	\\
        \hline
        DTLZ1				& 1.03e+2(2.34e+1)$\approx$ 	& 9.84e+1(2.04e+1) \\
        DTLZ2				& 1.57e-1(2.29e-2)$\approx$ 	& 1.62e-1(1.90e-2) \\
        DTLZ3				& 2.03e+2(2.42e+1)$\approx$ 	& 2.06e+2(2.13e+1) \\
	DTLZ4				& 4.91e-1(1.24e-1)$\approx$ 	& 5.20e-1(6.92e-2) \\	   			 	  	
	DTLZ5				& 1.18e-1(2.25e-2)$+$		& 1.11e-1(2.41e-2) \\	 	
	DTLZ6				& 1.29e+0(6.44e-1)$\approx$ 	& 1.36e+0(7.36e-1) \\		 	 	 
	DTLZ7				& 4.16e+0(2.54e+0)$\approx$ 	& 4.94e+0(2.31e+0)  \\		          	  
	\hline  			
	win/tie/loss 			& 0/7/0					& -/-/- \\
	\hline
	vs MOEA/D-EGO		& 4/2/1					& 4/2/1 \\
	vs 6 Comparisons		& 29/8/5					& 28/9/5 \\
	\hline
	\end{tabular}
\end{table}

\subsection{Performance on Extremely Expensive Multi-Objective Optimization: Result Figure}
\begin{figure*}[!t]
	\centerline{
	\includegraphics[width=3.2in]{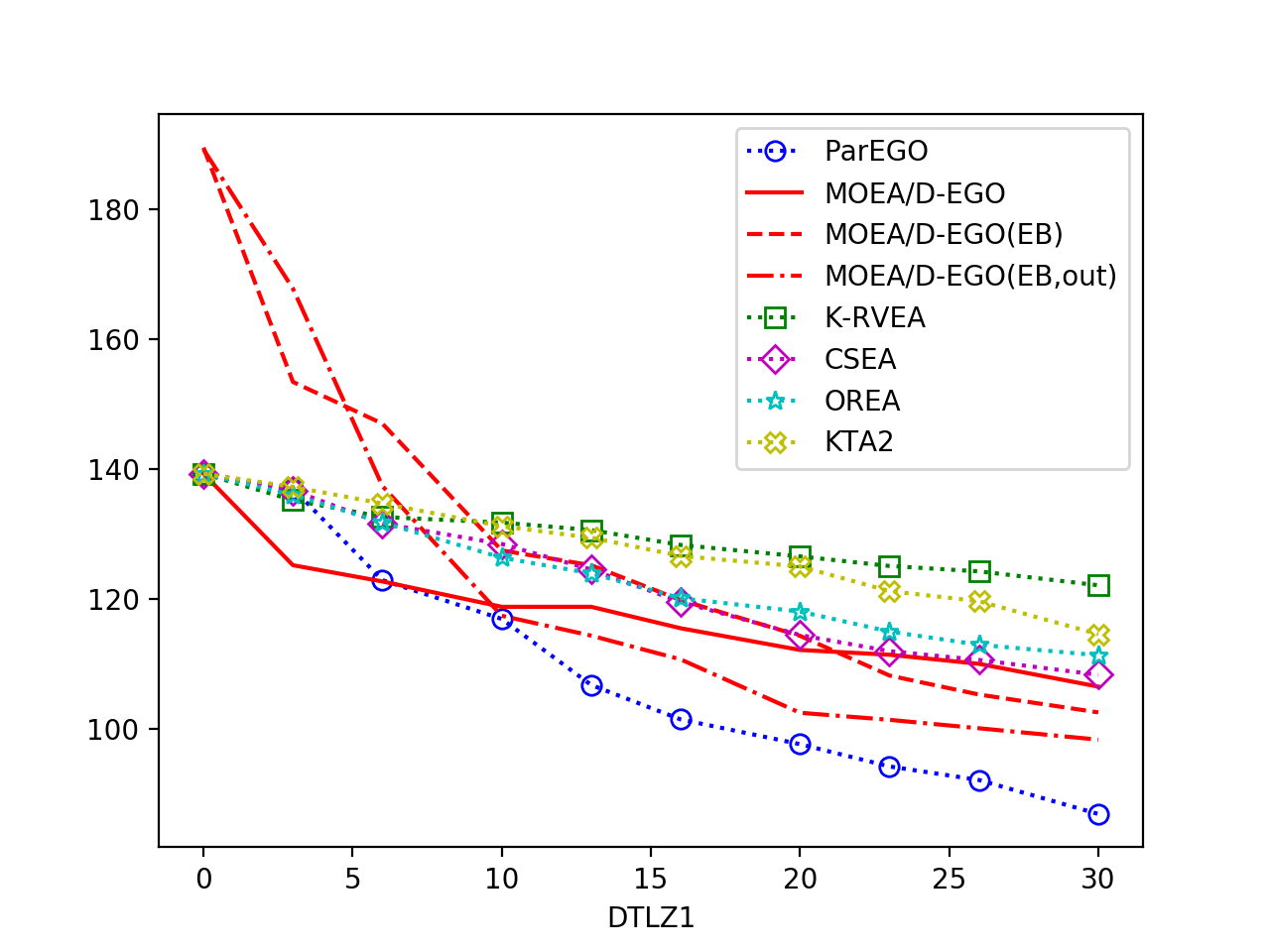}
	\includegraphics[width=3.2in]{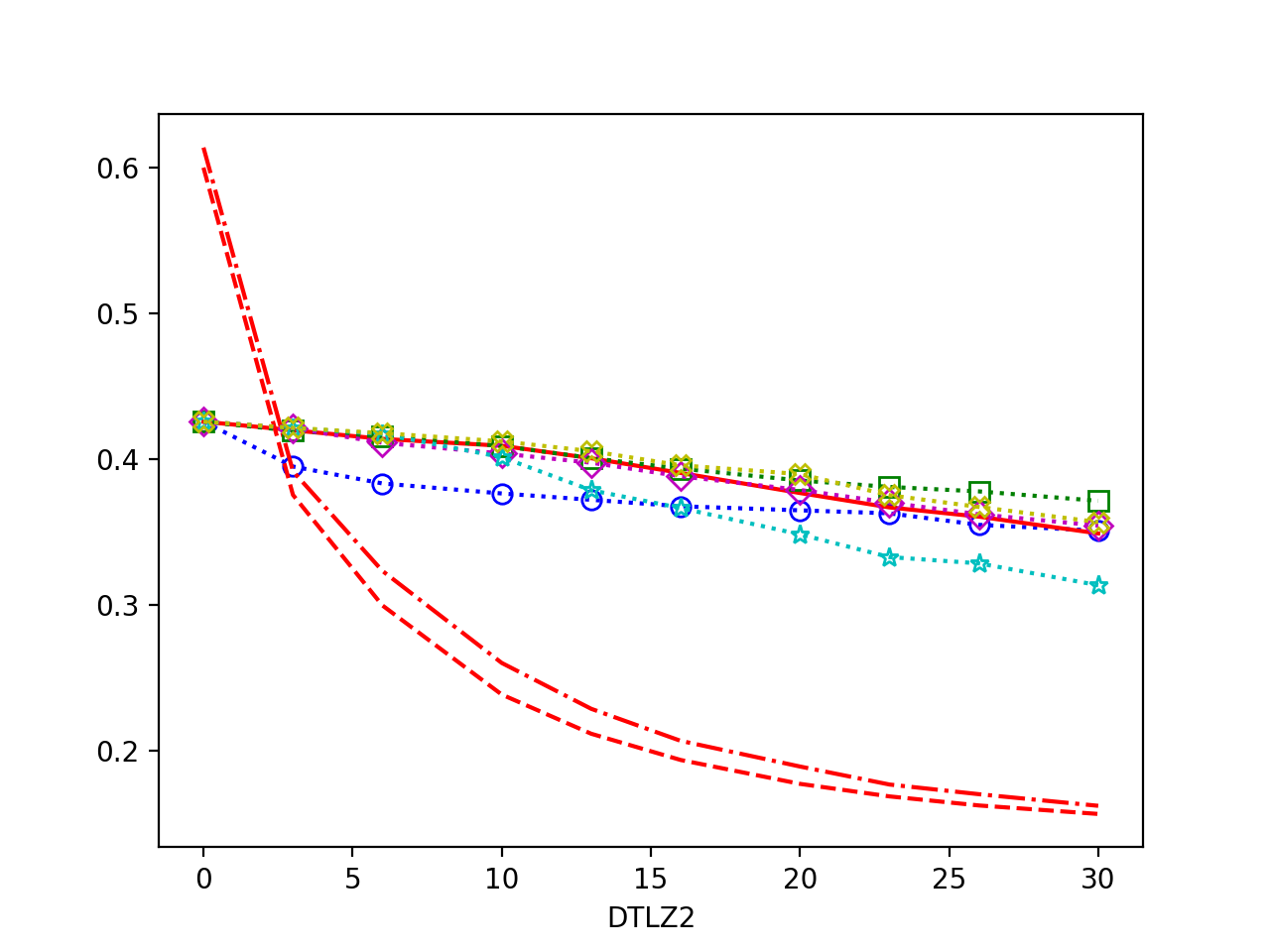}
	}
	\centerline{
	\includegraphics[width=3.2in]{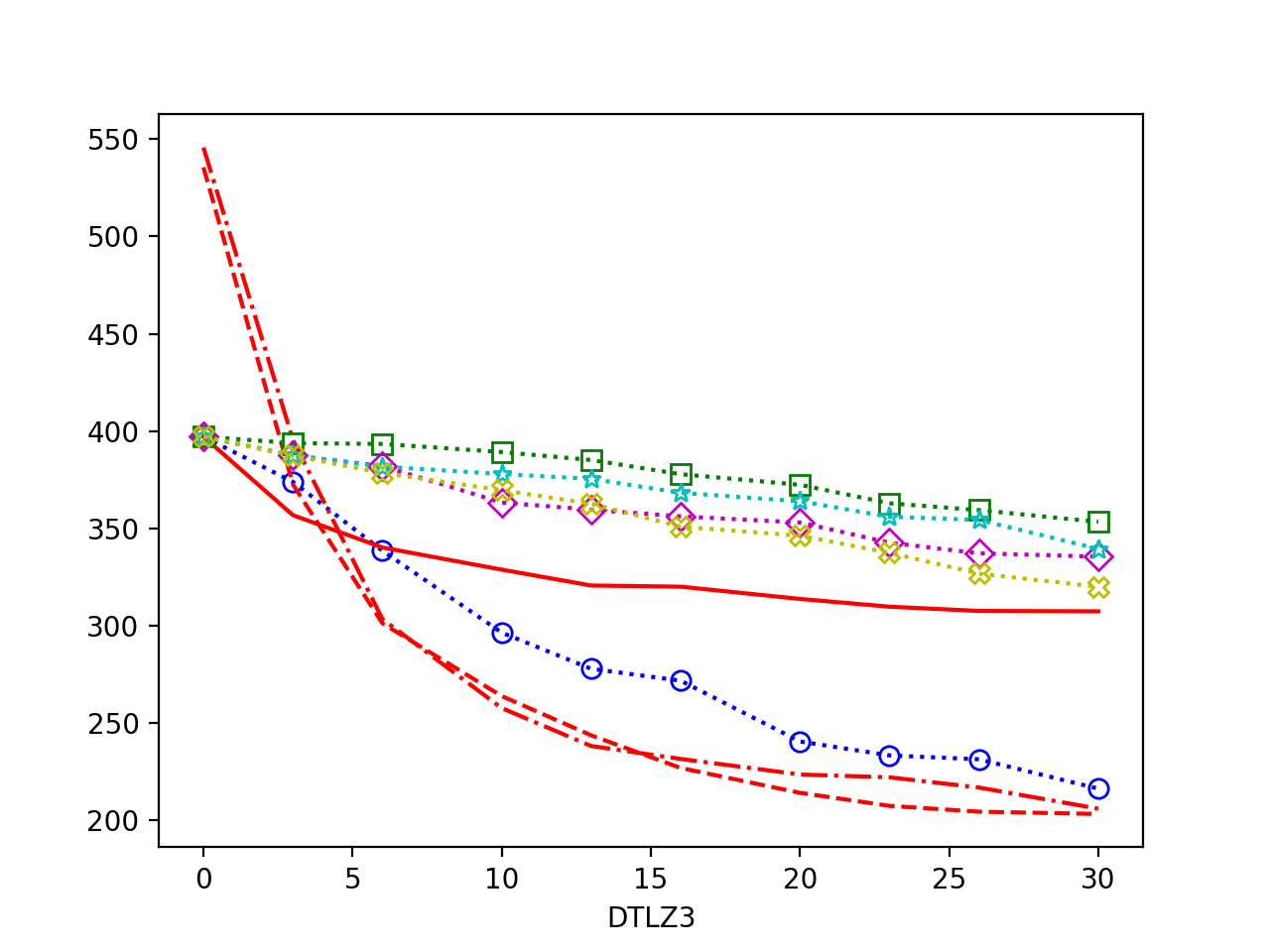} 
	\includegraphics[width=3.2in]{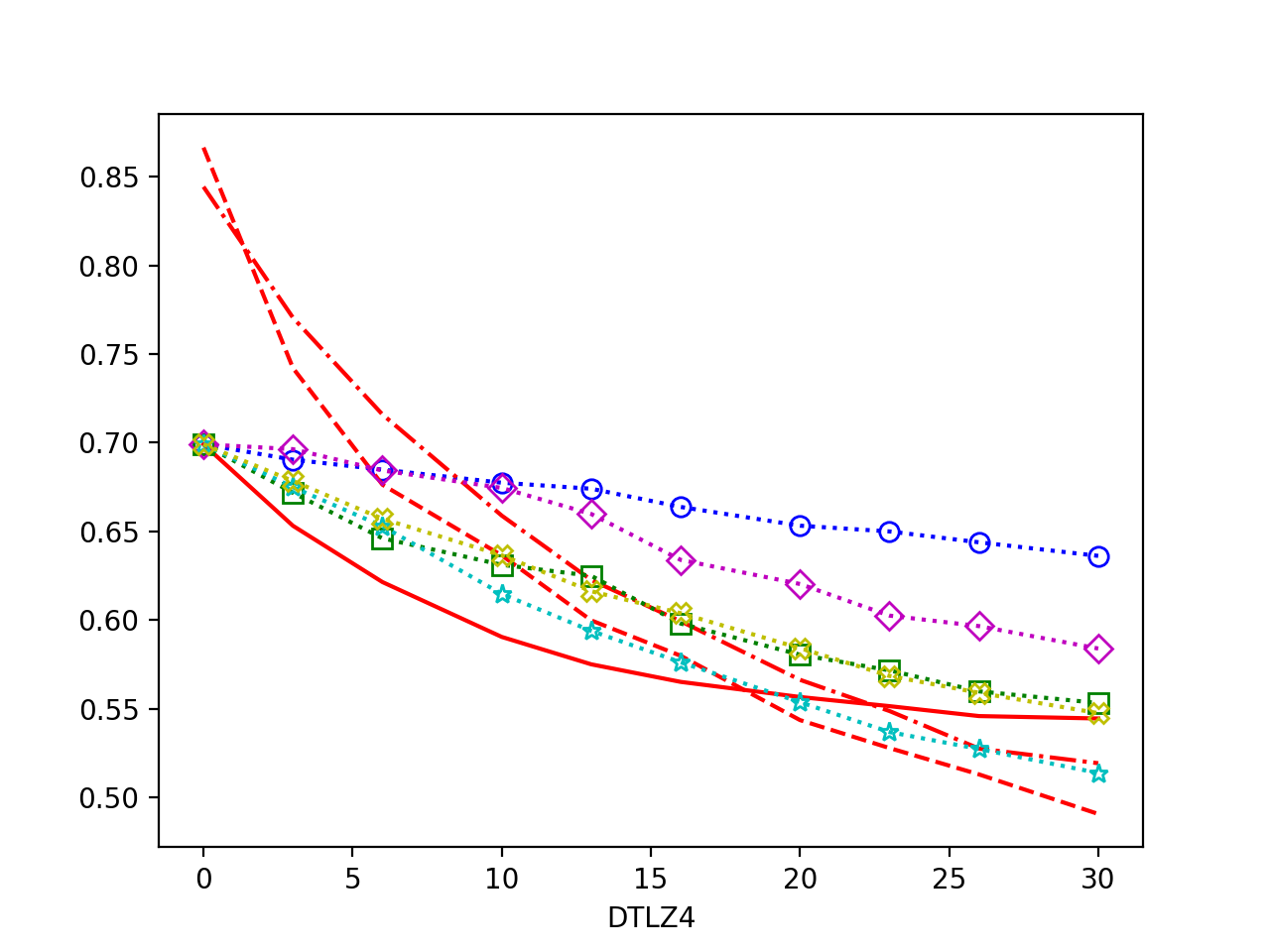} 
	}
	\centerline{
	\includegraphics[width=2.5in]{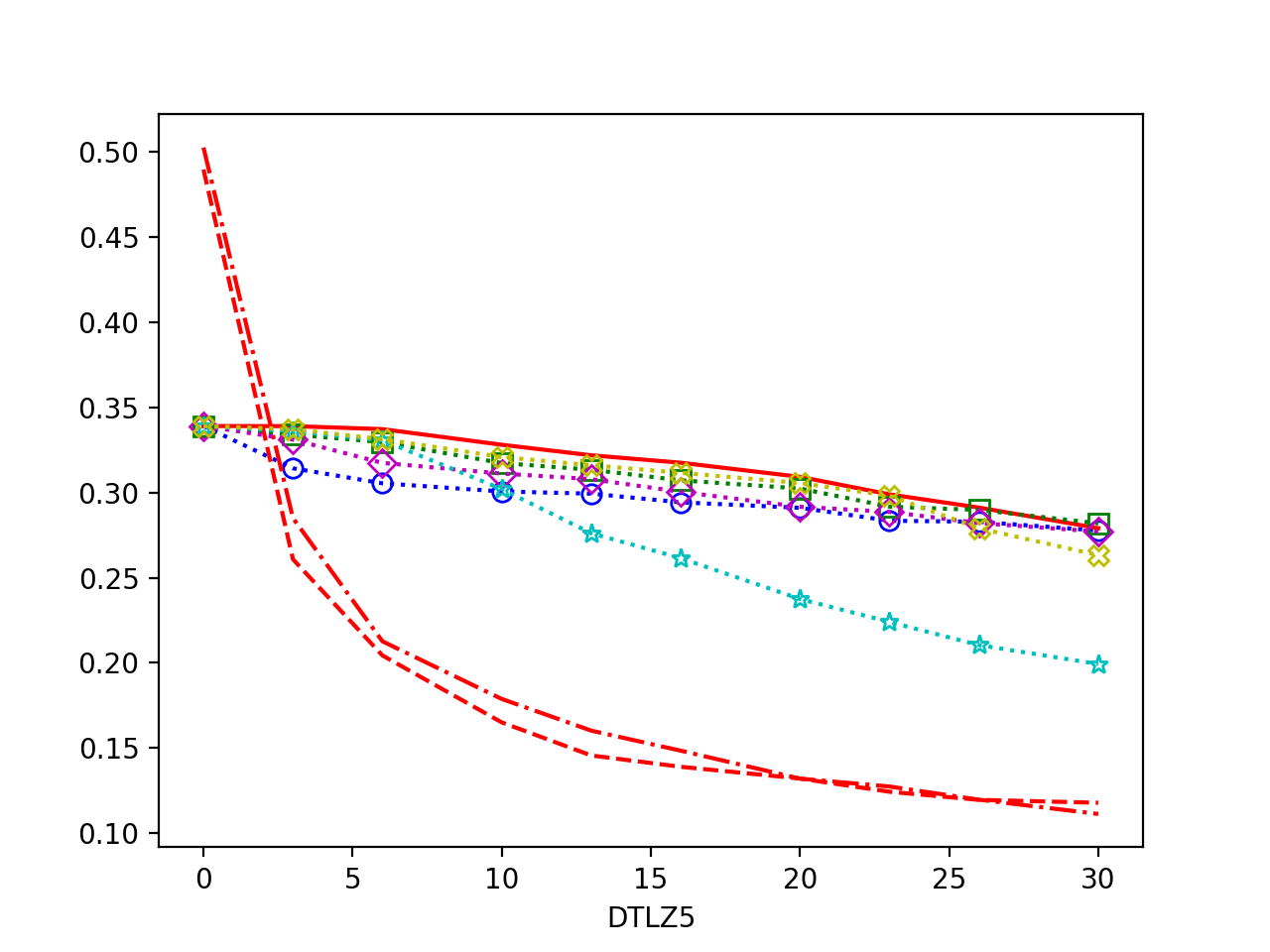} 
	\includegraphics[width=2.5in]{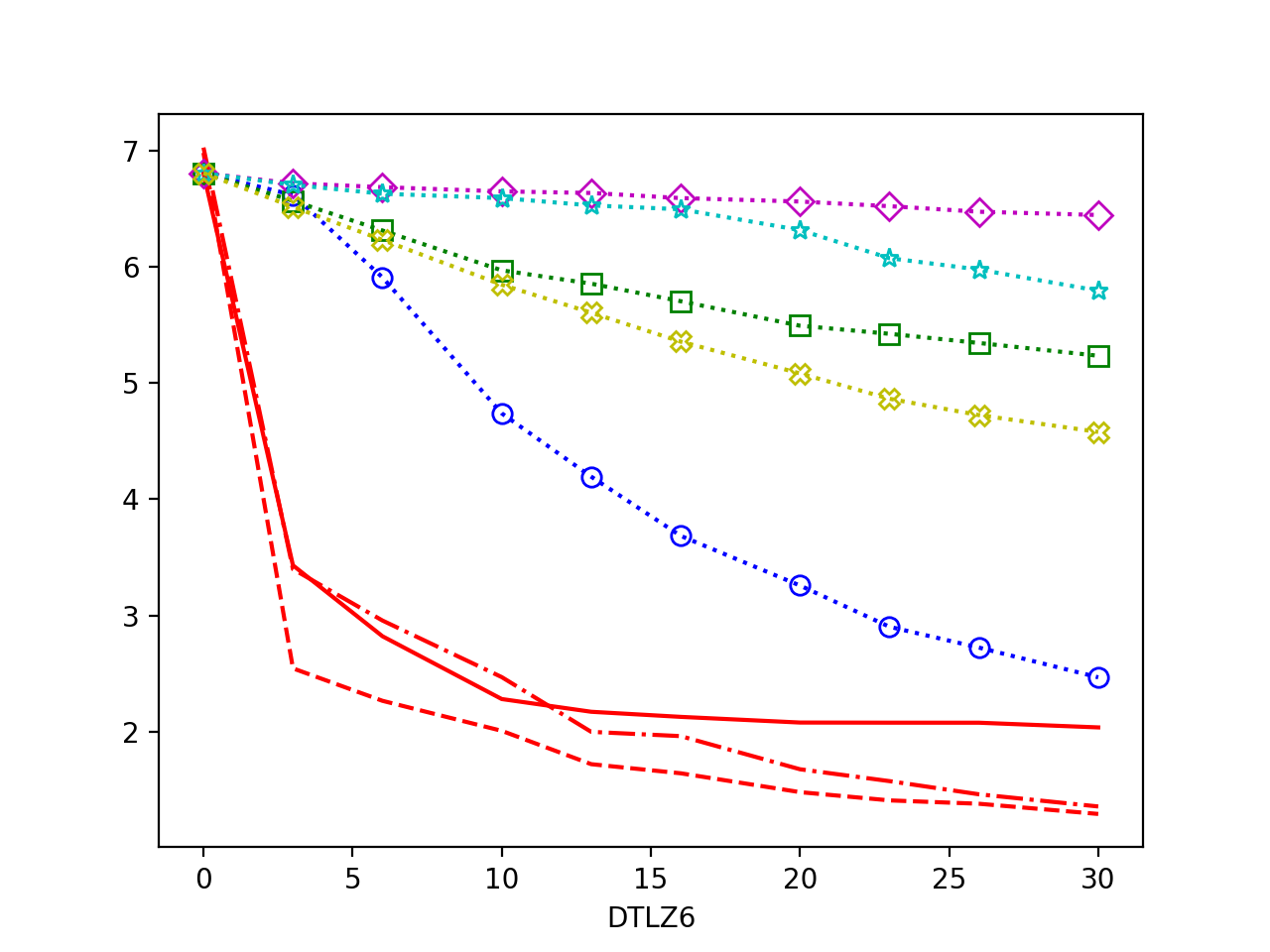} 
	\includegraphics[width=2.5in]{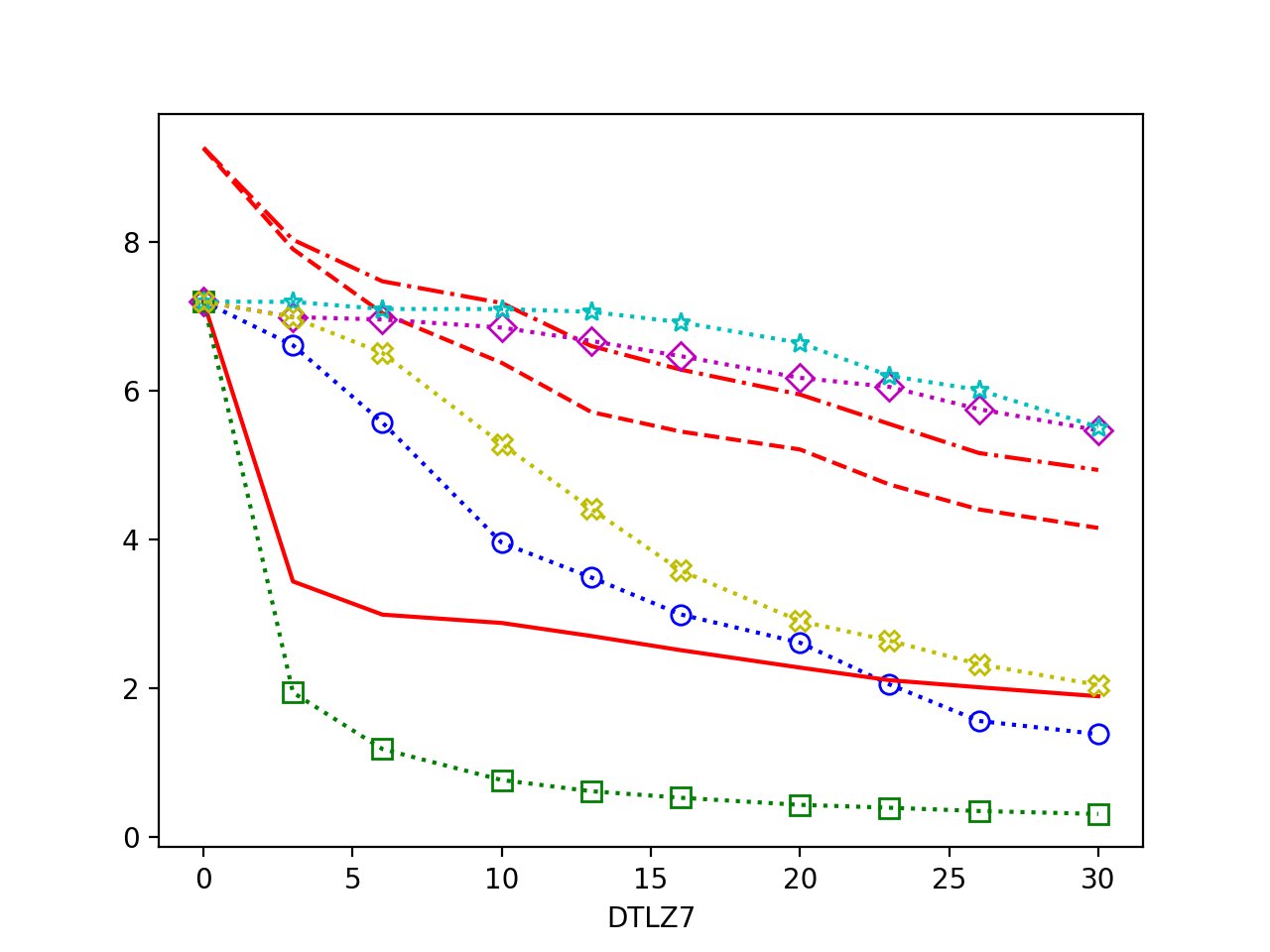} 
	}
	\caption[Average IGD+ values obtained from 30 runs on DTLZ functions. 30 evaluations for further optimization.]
	{Average IGD+ values obtained from 30 runs on DTLZ functions.
	MOEA/D-EGO(EB)s and comparison algorithms initialize their surrogates with 10, 60 samples, respectively. Extra 30 evaluations are allowed during the optimization process. 
	Note that 'EB out' indicates the target task is excluded from the range of related tasks during the MDKL meta-learning procedure.
	X-axis denotes the number of evaluations used after the surrogate initialization. 
	In comparison to MOEA/D-EGO, both MOEA/D-EGO(EB)s achieve smaller or competitive IGD+ values on all DTLZ testing functions except for DTLZ7, while 50 evaluations are saved with the assistance from related tasks. Moreover, MOEA/D-EGO(EB)s achieve the smallest IGD+ values on DTLZ2, DTLZ3, DTLZ4, DTLZ5 and DTLZ6.}
	\label{fig: DTLZ90}
\end{figure*}

Results for Sections \ref{sec: exp-DTLZ90} and \ref{sec: exp-out90} (in this supplementary material) are illustrated in Fig. \ref{fig: DTLZ90}. MOEA/D-EGO(EB, out) refers to the results obtained in the `out-of-range' situation.

\section{Summary of Experiments}
Our computational studies have demonstrated the following: 
First, we provide empirical evidence to show the effectiveness of learning experiences: The meta-learning of neural network parameters and base kernel parameters are essential to the modeling accuracy of a MDKL model. As a result, our MDKL model outperforms the compared meta-learning modeling and non-meta-learning modeling methods on both the engine fuel consumption regression task and the sinusoid function regression task.

Second, we demonstrate the main contribution of this work: 
In most situations, the proposed experience-based SAEA framework can assist regression-based SAEAs to reach competitive or even better optimization results, while the cost of surrogate initialization is only 1$d$ samples. 
Due to the effectiveness of saving evaluations, our SAEA framework is preferable to other SAEAs when solving problems within a very limited evaluation budget. 
Moreover, some empirical guidelines are concluded to help the application of our SAEA framework. 
For the influence of task similarity, we find that related tasks that are very similar to the target task are not necessary to the application of our approach. The influence of these similar tasks on the optimization performance is limited. Our experience-based SAEA framework can achieve competitive results without datasets from very similar related tasks. 
Besides, for the related tasks used for meta-learning, we have demonstrated that more useful experiences can be learned if more data points are sampled from related tasks.

Third, the effectiveness of our SAEA framework is validated on a real-world engine calibration problem. Competitive or better results are achieved on the objective and constraint functions, while 1$d$ samples are used to initialize surrogates. Therefore, our experience-based SAEA framework can also be applied to optimization scenarios such as single-objective optimization and constrained optimization.

\bibliographystyle{IEEEtran}
\bibliography{ExperienceBasedEA.bib}